\documentclass[journal]{IEEEtran}

\usepackage[left=0.625in,right=0.625in,top=0.75in,bottom=1in]{geometry}
\usepackage{ifpdf}
\usepackage{cite}
\usepackage{array}
\usepackage{dblfloatfix}
\usepackage{url}
\usepackage{makecell}
\usepackage{amsmath}
\usepackage{ragged2e}
\usepackage{graphicx}
\usepackage{verbatim}
\usepackage{enumitem}
\usepackage{float}
\usepackage{lipsum}
\usepackage{multicol, blindtext}
\usepackage{caption}
\usepackage{subcaption}
\usepackage{xcolor}
\definecolor{mypink2}{RGB}{0, 0, 255}
\definecolor{green}{RGB}{0, 128, 0}
\usepackage{multirow}
\usepackage{array}
\usepackage{tabularx}
\usepackage[acronym]{glossaries}
\usepackage[ruled,vlined]{algorithm2e}
\usepackage{algorithmic}
\usepackage[utf8]{inputenc}
\usepackage{tcolorbox}
\usepackage{xcolor}
\usepackage{soul}
\usepackage[normalem]{ulem}

\newcommand{\red}[1]{\textcolor{red}{[Hao: #1]}}

\usepackage{url}
\usepackage{hyperref}
\hypersetup{hidelinks}

\begin{document}

\title{\fontsize{18pt}{18pt}\selectfont Large Language Model-Assisted UAV Operations and Communications: A Multifaceted Survey and Tutorial}

\author{ Yousef~Emami,~\IEEEmembership{Senior Member,~IEEE,}
        Hao~Zhou,~\IEEEmembership{Member,~IEEE,}
        Radha~Reddy,
        Atefeh~Hajijamali Arani,
        Biliang Wang, 
        Kai~Li,~\IEEEmembership{Senior Member,~IEEE,}
        Luis~Almeida,~\IEEEmembership{Senior Member,~IEEE,}
        and~Zhu~Han,~\IEEEmembership{Fellow Member,~IEEE}
 
\thanks{Copyright (c) 2026 IEEE. Personal use of this material is permitted. However, permission to use this material for any other purposes must be obtained from the IEEE by sending a request to pubs-permissions@ieee.org.}     
 }

\maketitle

\begin{abstract}
Uncrewed Aerial Vehicles (UAVs) are widely deployed across diverse applications due to their mobility and agility. Recent advances in Large Language Models (LLMs) offer a transformative opportunity to enhance UAV intelligence beyond conventional optimization-based and learning-based approaches. By integrating LLMs into UAV systems, advanced environmental understanding, swarm coordination, mobility optimization, and high-level task reasoning can be achieved, thereby allowing more adaptive and context-aware aerial operations.
This survey systematically explores the intersection of LLMs and UAV technologies and proposes a unified framework that consolidates existing architectures, methodologies, and applications for UAVs. We first present a structured taxonomy of LLM adaptation techniques for UAVs, including pretraining, fine-tuning, Retrieval-Augmented Generation (RAG), and prompt engineering, as well as key reasoning capabilities such as Chain-of-Thought (CoT) and In-Context Learning (ICL).
We then examine LLM-assisted UAV communications and operations, covering navigation, mission planning, swarm control, safety, autonomy, and network management.
Subsequently, the survey discusses Multimodal LLMs (MLLMs) for human–swarm interaction, perception-driven navigation, and collaborative control. Finally, we discuss ethical considerations, including bias, transparency, accountability, and Human-in-the-Loop (HITL) strategies, and outline directions for future research. Overall, this work positions LLM-assisted UAVs as a foundation for intelligent and adaptive aerial systems.

\end{abstract}

\begin{IEEEkeywords}
Uncrewed Aerial Vehicles, Large Language Models, 6G Networks, Retrieval-Augmented Generation, Prompt Engineering, In-Context Learning, Chain-of-Thought.
\end{IEEEkeywords}

\IEEEpeerreviewmaketitle
\section{Introduction}

Uncrewed Aerial Vehicles (UAVs) have become a salient enabling technology across a wide range of applications, including environmental monitoring \cite{liu2022uav}, public safety \cite{shakoor2019role}, transportation \cite{menouar2017uav}, agriculture \cite{phang2023satellite}, and infrastructure inspection \cite{atefeh_ojvt24}. For example, UAVs can track environmental changes, support emergency response during disasters, assess crop health, deliver parcels, and assist with traffic management. Due to their flexibility and mobility, UAVs provide efficient and cost-effective solutions in situations where conventional ground-based systems are limited or infeasible \cite{10246260,atefeh_icc21,atefeh_access21}.
\par
From a communication perspective, UAVs are increasingly important for enhancing wireless networks. They can serve as flying base stations to expand network coverage, especially in areas with sparse or damaged infrastructure \cite{atefeh_iot22,atefeh_ojcs23}. UAVs can also operate as relays \cite{7192644}, \cite{pinto_almeida2022}, collect data from ground devices \cite{8854903}, and provide edge-computing capabilities in the air \cite{9793853}, thereby reducing communication latency. On the other hand, UAVs may operate as cellular users, requiring reliable, low-latency connectivity for safe navigation, control, and sensing \cite{azari_gcw20}. When multiple UAVs cooperate, they can form Flying Ad hoc Networks (FANETs) that self-organize and provide either extended range \cite{pinto_almeida2017} or communication services over large geographical areas or in environments without pre-existing infrastructure. These capabilities make UAVs a critical component of wireless backhauling, smart city applications, and future mobile communication systems \cite{8660516}.
\par
Advancements in cost efficiency, operational flexibility, and dynamic deployment capabilities are driving the rapid expansion of the UAV ecosystem. Modern UAV platforms are increasingly equipped with high-resolution cameras and precision sensors, enabling large-scale and cost-effective data acquisition. As a result, UAVs have become highly valuable for numerous civilian applications across diverse industries \cite{heiets2023future}. The global UAV market was valued at USD 36.41 billion in 2024 and is projected to 
reach USD 125.91 billion by 2032, with a Compound Annual Growth Rate (CAGR) of 17.3\% \cite{FortuneBusinessInsights2025AI}. This rapid market growth underscores the growing demand for intelligent UAV-assisted solutions.

In parallel with these developments, Artificial Intelligence (AI) has become a salient enabler for enhancing the autonomy and efficiency of UAV-assisted communication systems. AI techniques are widely used to optimize UAV mobility, resource allocation, and scheduling, allowing UAVs to adapt dynamically to both aerial and ground operations. These approaches support large-scale network management by enabling efficient and automated data collection. The adoption of AI is primarily driven by the inherent complexity and highly dynamic nature of UAV operating environments, where accurate analytical modeling is often infeasible. Additionally, UAV mobility creates high-dimensional state and action spaces, whereas limited sensing capabilities and intermittent connectivity result in incomplete or outdated information about ground devices, including battery levels, energy availability, and channel conditions. These challenges make data-driven and learning-based solutions particularly well-suited to UAV network optimization \cite{FortuneBusinessInsights2025AI}.

Recently, AI has advanced significantly with the emergence of Large Language Models (LLMs), which provide enhanced capabilities in language understanding, generation, and reasoning. Building on decades of progress in natural language processing and Machine Learning (ML), LLMs have become a major driver of recent developments in AI-assisted systems. Their ability to interpret complex instructions and perform multi-step reasoning has generated substantial interest across a wide range of application domains. In UAV systems, LLMs offer new opportunities to automate complex decision-making processes, enhance situational awareness, and support intelligent network operations \cite{kaleem2024emerging}. However, deploying LLM-assisted solutions in safety- and mission-critical environments introduces significant ethical and operational challenges, including data privacy and security, bias and fairness, and accountability and transparency. Addressing these concerns is essential to ensure the trustworthiness and reliable integration of LLMs.

Moreover, while state-of-the-art LLMs primarily operate on textual data, UAV-assisted networks inherently generate heterogeneous information, including wireless signals, visual sensing data, and contextual metadata. This modality mismatch has motivated the development of Multimodal LLMs (MLLMs) and Vision-Language Models (VLMs) that can jointly process and reason over diverse data sources \cite{han2025multimodal}. This multimodal intelligence is particularly relevant to UAVs, where effective decision-making often depends on integrating communication, sensing, and environmental information.

\subsection{Why Combining UAVs and LLMs Matters}

The integration of UAVs with LLMs introduces a new level of intelligence that conventional control and optimization methods struggle to achieve. Optimization has long been central to UAV communications, such as resource allocation and trajectory design. These problems typically require selecting optimal solutions from discrete decision spaces to maximize objectives, such as throughput, under strict operational constraints. Conventional optimization techniques, such as convex optimization and metaheuristic algorithms, lack sufficient adaptability in highly dynamic UAV environments. Although ML and Deep Reinforcement Learning (DRL) techniques offer greater flexibility, they often exhibit limited generalization and a persistent simulation-to-reality gap, hindering their reliable deployment in real-world UAV scenarios.

Compared with conventional ML-based optimization methods, LLMs introduce a fundamentally different paradigm centered on semantic reasoning rather than parameter tuning alone. While traditional ML and DRL approaches optimize numerical objectives through repeated interactions with simulated environments, LLMs operate at a higher level of abstraction, enabling structured reasoning over heterogeneous inputs and mission-level objectives. This capability allows LLMs to function not merely as optimizers but as cognitive coordinators within UAV systems.
Specifically, in UAV communications and networking scenarios, LLMs can interpret network logs, topology information, channel conditions, and mission constraints in natural-language form, thereby transforming raw telemetry into structured situational awareness. Instead of solving isolated optimization subproblems (e.g., power allocation or trajectory refinement), LLMs can synthesize cross-layer information to guide adaptive spectrum management, cooperative relay strategies, and swarm-level resource coordination. By reasoning over operational rules, safety constraints, and mission priorities, LLMs can generate context-aware decision strategies that complement lower-level control modules.

Beyond communication optimization, LLMs enhance UAV autonomy by bridging perception, planning, and execution. Through multimodal integration, LLMs can align sensor data, map representations, and textual mission descriptions into a unified reasoning framework. This enables dynamic task decomposition, contingency planning, and collaborative swarm coordination under uncertain or partially observable conditions. Rather than relying solely on fixed reward functions, LLM-driven systems can adapt strategies to evolving mission semantics and operator intent.
In addition, this reduces the dependence on handcrafted rule sets and specialized pipelines and mitigates the need for multiple task-specific models. By unifying diverse data modalities—natural-language commands, telemetry streams, network metrics, and environmental semantics—LLMs provide a coherent operational layer that supports complex, multi-objective UAV missions.

A defining feature of LLMs is In-Context Learning (ICL), which enables real-time adaptation to user requirements through feedback alone, without the computational overhead of retraining. This adaptability, combined with inherent scalability, allows LLMs to handle the combinatorial complexity of multi-UAV optimization by decomposing high-level, multi-objective missions into coherent sub-tasks and executable strategies. Furthermore, the rich semantic knowledge embedded in LLMs supports decision-making that closely aligns with human intuition and established operational doctrines.
These reasoning capabilities also improve transparency and trust by enabling UAV systems to generate natural-language explanations of their actions. This explainability enables more intuitive communication with human operators, reduces cognitive and operational burdens, and allows non-expert users to control UAVs effectively. Ongoing advances in communication technologies, such as 6G and satellite internet, provide low-latency, high-bandwidth connectivity required to support real-time LLM-assisted UAV operations, thereby accelerating the transition toward intelligent, autonomous aerial systems.


In the following, we present a unified system-level abstraction that explains how LLMs interact holistically with UAV communication, control, and decision-making. This is {\em a macro-level methodological framework for understanding LLM–UAV integration that is organized into three layers}: 
\begin{itemize}
    \item At the highest layer, the cognitive and decision-making layer assigns LLMs the role of meta-controller, responsible for high-level reasoning and mission orchestration. Rather than operating on raw sensor data, the LLM processes symbolic and semantic representations of mission goals, environmental context, and system constraints. This layer translates UAV intent and abstract objectives into structured task descriptions, enabling mission planning, task decomposition, constraint reasoning, and adaptive replanning in dynamic environments. 
    \item The perception layer mediates between cognition and execution by converting multimodal sensor inputs into semantically meaningful descriptions. Data from vision, LiDAR, radar, and spectrum sensors are fused into higher-level concepts such as objects, activities, behaviors, and inferred intent. LLMs, often combined with MLLMs, provide contextual interpretation, allowing the UAV to understand not only what is observed but also how it relates to mission objectives. 
    \item The control and communication layer, at the bottom, translates high-level plans and semantic decisions into executable actions, such as control commands, flight trajectories, and communication policies. Execution outcomes and system states are abstracted and relayed to higher layers, forming a closed-loop learning and adaptation process. 
\end{itemize}
    This layered abstraction unifies the diverse LLM-assisted methods surveyed in this paper by framing LLMs as cross-layer cognitive integrators rather than isolated components. It clarifies how LLMs complement traditional UAV subsystems, enabling scalable autonomy, semantic reasoning, and adaptive decision-making in communication, control, and mission planning.
    

\subsection{Main Contributions}

This paper presents a comprehensive investigation of LLM-UAV interactions. It systematically examines how LLM-assisted techniques, such as Retrieval-Augmented Generation (RAG) and prompt engineering, are fundamentally transforming UAV operations by enhancing decision-making, enabling adaptive path planning, and supporting real-time human–UAV interaction. 
In addition, we explore the paradigm shift from text-only LLMs to MLLMs, analyzing their architectures, training methodologies, and emerging applications in vision–language navigation, swarm coordination, and high-level mission planning. 
Beyond technical advancements, we critically assess the ethical implications of integrating LLMs into UAV operations and communications, and identify directions for future research.
Overall, this work serves as a comprehensive technical survey and tutorial, as well as an essential ethical guide for the evolving landscape of LLM-assisted autonomous UAV systems.

The main contributions of this survey are the following:

\begin{itemize}
    \item \textbf{Comprehensive synthesis and unified framework} that systematically maps the convergence of LLMs and UAV systems across operational, communication, and ethical dimensions. By consolidating existing fragmented research, it provides a unified framework and a clear overview of existing architectures, methodologies, and real-world applications. In addition, we introduce a structured taxonomy of LLM approaches explicitly tailored to UAV challenges, encompassing pretraining, fine-tuning, RAG, and prompt engineering. This taxonomy clarifies the current research landscape and highlights promising directions for future development at the intersection of LLMs and UAVs.

    \item \textbf{In-depth discussion of MLLMs} as a transformative enabler for next-generation UAV systems, detailing their core architectures, training paradigms, and emerging capabilities such as Multimodal Chain of Thought (MCoT) reasoning and Multimodal ICL (M-ICL). We further demonstrate that MLLMs enable vision–language navigation and facilitate coordinated swarm control, underscoring their growing importance for complex perception, reasoning, and decision-making tasks in autonomous UAVs.

    \item \textbf{Rigorous analysis of the ethical challenges} associated with deploying  LLM-assisted UAV systems (LAUS), including risks related to bias and fairness, transparency and accountability in autonomous decision-making, and the environmental impact of LLMs. Together, these analyses establish a practical foundation for building trustworthiness, responsibility, and resilience in LLM-assisted UAV communications.
\end{itemize}

\subsection{Survey of Surveys}

Table~\ref{tab:llm-uav-surveys} summarizes key surveys on integrating LLMs with UAVs. Early comprehensive research by Javaid et al.~\cite{10643253} examines how LLMs can enhance UAV autonomy, intelligence, and operational efficiency. Their survey reviews a wide range of notable LLM architectures, including Bidirectional Encoder Representations from Transformers (BERT) \cite{devlin2019bert}, Generative Pretrained Transformers (GPT), T5, XLNet, and evaluates their applicability to various UAV tasks such as natural language-based control, real-time data analysis, and mission planning. This work lays the foundation for understanding the technical potential of LLM-aided UAVs.

\begin{table*}[h]
\centering
\renewcommand{\arraystretch}{1.25}
\caption{ A summary of existing surveys on the integration of LLMs and MLLMs in UAV Operation and Communication.}
\label{tab:llm-uav-surveys}
\begin{tabular}{|m{0.8cm}|m{6.5cm}|m{9.5cm}|}
\hline
\textbf{Surveys} & \textbf{Focus \& Scope} & \textbf{Key Findings / Application Categories} \\
\hline
\cite{10643253} & A broad overview of integrating various LLMs (BERT, GPT, T5, etc.) with UAVs to enhance autonomous systems. & Applications in natural language control, real-time data analysis, and mission planning for domains like surveillance and disaster response. \\
\hline
\cite{cidjeu2025uav} & Analyzes 17 papers to map the research landscape (2021–2024). & \textbf{Applications:} UAV Optimization (35\%), Adaptation (40\%), Communication (25\%). \textbf{Models:} GPT most used (58\%). Highlights challenges, including hardware limitations, latency, security, and ethical alignment. \\
\hline
\cite{yumeng2025empowering} & Focuses on revolutionizing UAV-assisted wireless communication, information perception, and autonomous control. & Enhances aerial networking, edge computing, multimodal data fusion, scene understanding, and path planning. \\
\hline
\cite{kheddar2025recent} & Comprehensively surveys Transformer-based models (ViTs, STTs, etc.) and LLMs for UAV systems. & Categorizes architectures for object detection, tracking, navigation, and mission planning. Highlights LLMs for high-level reasoning and interaction. \\
\hline
\cite{chen2025large} & Analyzes 74 papers and 56 projects across 9 core tasks in the UAV workflow. & \textbf{Academia:} Broad, theoretical focus (e.g., multi-UAV planning). \textbf{Industry:} Practical focus on flight control, single-UAV planning, and human–machine interaction. \\
\hline
\cite{li2025applications} & A systematic review of LLM and MLLM integration in autonomous driving systems. & Explores applications in perception, prediction, planning, decision-making, multitasking processing, and HMI. Discusses multimodal fusion, prompt engineering, and distillation, outlining challenges and future directions. \\
\hline
\cite{cui2024survey} & Systematic investigation of MLLMs for autonomous driving and mapping systems. & Reviews MLLM tools, datasets, and benchmarks; summarizes results from the LLVM-AD workshop; identifies major challenges for future adoption in driving systems. \\
\hline
This survey & LLMs and MLLMs that enhance UAV autonomy in navigation, communication, and swarm coordination, using techniques such as fine-tuning, RAG, and prompt engineering. The survey discusses ethical issues such as bias and transparency. Practical and future research directions are provided. It positions LLM-assisted UAVs as a transformative, intelligent aerial framework.&LLMs enhance UAV navigation, path planning, and real-time swarm coordination, surpassing traditional optimization methods. In communications, they optimize spectrum management, resource allocation, and the integration of 5G/6G networks. MLLMs enable multimodal tasks like search and rescue by fusing visual, sensor, and language data. The technology also addresses ethical deployment through safety verification and human oversight frameworks.\\
\hline
\end{tabular}
\end{table*}

Building on this foundation, Cidjeu et al.~\cite{cidjeu2025uav} present a structured mapping of research published between 2021 and 2024, categorizing 17 representative studies into UAV optimization, adaptation, and communication. Their analysis reveals that GPT-based models currently dominate integration efforts, highlighting ongoing challenges related to onboard computational constraints, latency, security, and system robustness. Similarly, Yumeng et al.~\cite{yumeng2025empowering} examine the transformative role of LLMs in UAV-assisted wireless communications, multimodal perception, and autonomous control. Their survey demonstrates how LLMs support edge computing, aerial networking, and trajectory planning through enhanced reasoning and language understanding. They also emphasize the importance of addressing issues such as model adaptation, reliability, and real-time deployment in dynamic environments.
\par
Complementing these LLM-assisted studies, Kheddar et al.~\cite{kheddar2025recent} broaden the scope by surveying transformer-based models for UAV applications, ranging from hybrid Convolutional Neural Network (CNN)-Transformer architectures to vision and spatio-temporal transformers. Their taxonomy includes contributions in object detection, tracking, and mission planning, and highlights the growing importance of LLMs for high-level reasoning and human–UAV interaction. However, they also identify critical challenges, including data scarcity, multimodal sensor fusion, and the difficulty of meeting real-time operational requirements.
\par
A more extensive empirical study by Chen et al.~\cite{chen2025large} analyzed 74 academic publications and 56 industrial projects to identify nine core LLM-assisted tasks spanning the entire UAV workflow, from information input to result feedback. Their comparative study reveals a clear gap between academic research, which often emphasizes complex multi-UAV planning and coordination, and industrial practice, which prioritizes practical functionalities such as flight control and human–machine interaction. Feedback from 52 practitioners further highlights concerns about safety, performance, and cost, suggesting that LLMs are currently better suited to supportive and decision-assistance roles rather than direct control in safety-critical UAV systems. In a concise follow-up, Cidjeu et al.~\cite{cidjeu2025uav} reiterate the importance of natural language interfaces, multimodal understanding, and dynamic code generation, and call for a deeper investigation into multi-UAV collaboration and LLM-assisted autonomous decision-making.
\par
Although these surveys primarily focus on UAVs, adjacent domains provide complementary insights. In autonomous driving, Li et al.~\cite{li2025applications} present a systematic review of the integration of LLMs and MLLMs across perception, prediction, planning, decision-making, and human–machine interaction. Their discussion of multimodal fusion, knowledge distillation, and fine-tuning strategies closely aligns with the challenges encountered in UAV systems. Similarly, Cui et al.~\cite{cui2024survey} review MLLMs for autonomous driving and mapping, summarize datasets and benchmarks, and identify emerging research trends that require coordinated academic and industrial efforts.
\par
Unlike existing surveys, our work provides a focused, in-depth examination of UAV communications, emphasizing how LLMs can enhance communication reliability, autonomy, and decision-making in this critical field. We systematically analyze the LLM functionalities most relevant to UAV communication systems and provide detailed insights into their practical adoption. We also explore the emerging role of MLLMs in integrating visual, sensor, and textual data to enable advanced communication-aware UAV applications. Beyond technical aspects, we analyze ethical and societal implications, including safety, privacy, accountability, and potential misuse of LAUS, and discuss emerging trends and open research directions, positioning our survey as a timely and distinctive contribution to the evolving field of intelligent UAV communications.

\begin{figure*}[!thbp]
  \centering
  \includegraphics[width=0.95\textwidth]{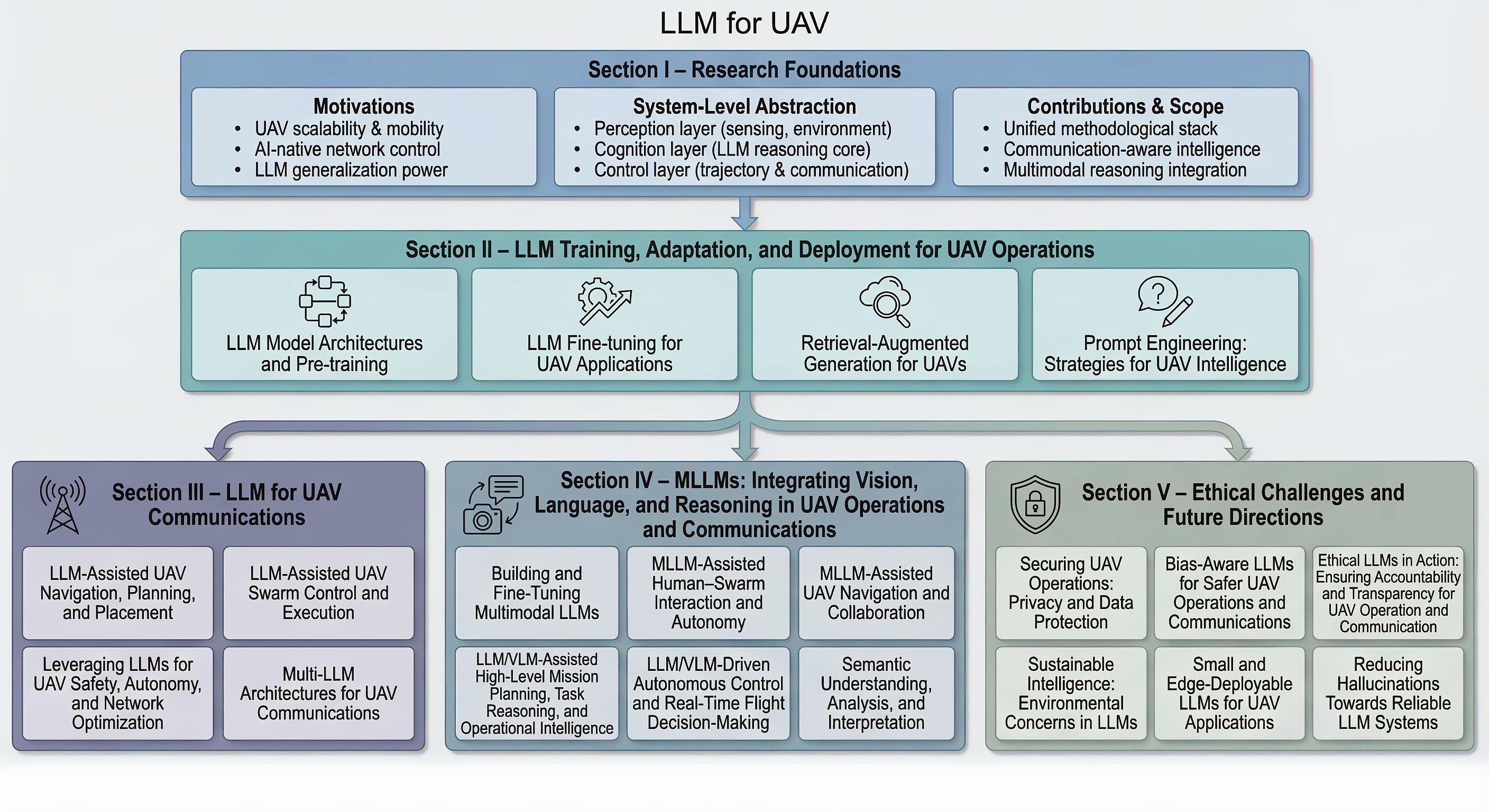} 
  \caption{Overall organization of the survey, illustrating the research foundations, LLM adaptation stack, core application domains (LLM-enabled communications and MLLM-driven intelligence), and responsible and future intelligence considerations for UAV systems.}  \label{fig:all20}
\end{figure*}

\subsection{Organization of This Survey}

The remainder of this paper is systematically structured as follows.  Fig. \ref{fig:all20} provides the general organization of the survey. Section~\ref{sec3} details the methodologies for LLM pre-training, adaptation (including fine-tuning, RAG, and prompt engineering), and deployment strategies tailored for UAV operations. Section~\ref{sec4} explores the application of LLMs specifically in UAV communications, covering navigation, swarm control, and network optimization. Section~\ref{sec5} examines the transformative role of MLLMs, detailing their architectures and their application in Vision-Language Navigation (VLN) and human-swarm interaction. Section~\ref{sec6} critically examines the associated ethical implications and future research directions, and finally Section~\ref{sec9} concludes the survey.

\section{LLM Training, Adaptation, and Deployment for UAV Operations} \label{sec3}

This section presents LLM model architectures and investigates their training, adaptation, and deployment for UAV operations. Training LLMs from scratch enables strong general capabilities, but is elusive for UAV applications due to prohibitive computational, energy, and time costs. 
Fine-tuning pretrained LLMs is an efficient approach to adapt model architectures to UAV-specific tasks, such as mobility and swarm optimization. However, it requires careful control over overfitting and the use of labeled data. RAG further improves LLM deployment by integrating external knowledge sources with LLM architectures, enabling access to up-to-date operational data and airspace regulations without retraining. Prompt engineering provides a lightweight approach to leveraging existing LLM architectures for rapid deployment and human–UAV interaction, enabling flexible and resource-efficient task optimization.

\subsection{LLM Model Architectures and Pre-training}

Large language models can be broadly categorized into three architectural paradigms: encoder-only, encoder–decoder, and decoder-only.
Encoder-only models (e.g., BERT) rely on bidirectional self-attention to produce deep contextual representations, making them particularly effective for language understanding and classification tasks. Encoder–decoder models (e.g., T5, BART) separate input encoding and output generation with cross-attention, enabling efficient handling of tasks that require both comprehension and structured generation. Decoder-only models (e.g., GPT-style architectures) adopt causal self-attention for autoregressive text generation, excelling in open-ended reasoning and interactive language production.
Their differences become clear in UAV applications. For example, in UAV log analysis, a BERT-based encoder-only model has been used for named entity recognition (NER) to extract forensic entities from DJI flight logs, achieving high F1 performance. This reflects its strength in structured understanding. In contrast, encoder–decoder models such as T5 are better suited for tasks like translating natural-language commands into structured UAV instructions or summarizing sensor data. Decoder-only models further extend this capability by enabling interactive UAV control, in which systems such as ChatGPT are integrated with PX4/Gazebo to generate executable flight commands via prompt-based reasoning.

Overall, the architectural choice determines whether the model primarily excels at understanding, understanding plus structured generation, or open-ended generative interaction, which directly shapes its role in UAV intelligence systems.

Pretraining equips LLMs with general language understanding and reasoning, primarily through next-token prediction on large-scale text corpora. This process enables models to capture long-range dependencies and acquire broad world knowledge before being adapted to specific domains.
In UAV communications, pretraining typically combines general corpora (e.g., Wikipedia and technical web content) with UAV-specific sources such as 3GPP standards, RFC documents, academic papers, protocol specifications, and network logs. Incorporating these domain-relevant materials improves the model’s understanding of UAV architectures, communication protocols, and optimization tasks, thereby supporting more reliable UAV-oriented reasoning.
However, training a large model from scratch for UAV applications is extremely resource-intensive. UAV data (e.g., flight logs, communication traces, and sensor records) are highly specialized and computationally expensive to process at scale. For example, training frontier-scale models such as LLaMA 3.1 (405B) requires massive GPU resources and prohibitive cost. As a result, full pretraining is typically impractical for UAV systems, making fine-tuning of pretrained models the more feasible and cost-effective approach for domain adaptation.

\subsection{LLM Fine-tuning for UAV Applications}

Fine-tuning adapts a pretrained LLM to domain-specific tasks using relatively small datasets, significantly reducing data and computational requirements compared to training from scratch. Using representations learned during large-scale pretraining, fine-tuning enables efficient specialization for UAV networks, improving task performance while maintaining generalization. This makes it a practical approach for deploying LLMs under UAV-specific constraints, including limited computational resources and dynamic operating environments.
\par
There are three main fine-tuning paradigms. Unsupervised fine-tuning adapts models using unlabeled domain text, improving domain familiarity but offering limited task precision. Supervised Fine-Tuning (SFT) relies on labeled datasets and often yields strong performance, though annotation costs can be prohibitive in UAV communications. Instruction fine-tuning, commonly implemented through prompt engineering, guides models using natural-language instructions and reduces labeling requirements, although performance is highly dependent on prompt design. Broadly, the fine-tuning plays a critical role in adapting pretrained LLMs to UAV tasks by balancing performance, efficiency, and deployment feasibility \cite{parthasarathy2024ultimate}.

\begin{figure}[!thbp]
  \centering
  \includegraphics[width=0.4\textwidth]{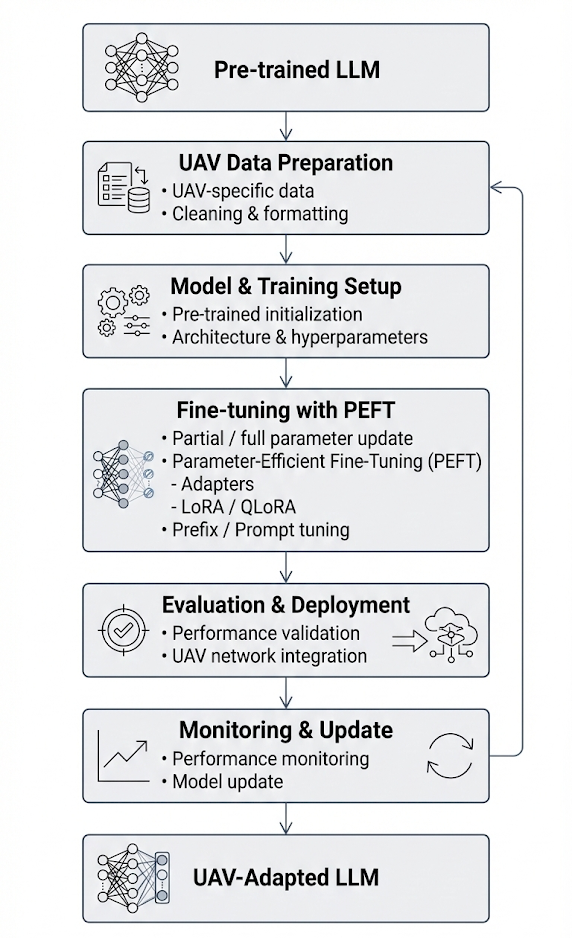} 
  \caption{LLM fine-tuning pipeline for UAV applications, showing the progression from a pretrained language model through UAV-specific data preparation, model and training setup, parameter-efficient fine-tuning (e.g., adapters, LoRA/QLoRA, and prompt tuning), followed by evaluation, deployment, continuous monitoring, and updates to produce a UAV-adapted LLM.
}
  \label{fig:finetune}
\end{figure}

As shown in Fig.~\ref{fig:finetune}, fine-tuning for UAV applications typically follows a structured pipeline encompassing UAV-specific data preparation, model initialization, training environment configuration, parameter updates, evaluation, deployment, and long-term monitoring. Model parameters may be partially or fully updated, depending on resource constraints, using Parameter-Efficient Fine-tuning (PEFT) \cite{xu2023parameter} to reduce overhead. 
Here, the PEFT adapts large pretrained models by freezing most parameters and updating only a small subset, such as adapters, low-rank matrices, or prompt embeddings. This strategy substantially reduces training cost, memory footprint, and convergence time while preserving pretrained knowledge and mitigating catastrophic forgetting. Representative PEFT methods include Adapters, Low-Rank Adaptation (LoRA), Quantized Low-Rank Adaptation  (QLoRA), Prefix-tuning, Prompt-tuning, and P-tuning, enabling efficient customization of models such as GPT-3, BERT, and LLaMA for UAV applications with limited resources \cite{belcic2024parameter}.
\par

Overall, fine-tuning offers a practical alternative to full pre-training by reducing computational and temporal costs, though it introduces risks such as overfitting and dependence on labeled data. With techniques such as PEFT, even models with tens to hundreds of billions of parameters can be adapted on modest hardware \cite{singh2024study}. In UAV networks, fine-tuning supports task-specific adaptation for applications such as troubleshooting and configuration; however, rapidly evolving domain knowledge often necessitates combining fine-tuning with RAG to maintain up-to-date and context-aware performance.


\begin{table*}[t]
\renewcommand{\arraystretch}{1.25}
\centering
\caption{Summary of RAG frameworks applied to UAV operation and communication systems, covering decision-making, network optimization, autonomous landing, edge computing, and evaluation methodologies.}
\label{tab:rag_uav_summary}
\resizebox{\textwidth}{!}{%
\begin{tabular}{|
m{1cm}|
m{3.5cm}|
m{3.5cm}|
m{5.5cm}|
m{4.5cm}|}
\hline
\textbf{Reference} &
\textbf{Proposed Framework} &
\textbf{Primary Application Domain} &
\textbf{Core Technical Approach} &
\textbf{Key Performance Results} \\
\hline
\cite{sezgin2025llm} &
Hybrid LLM--RAG Decision-Making System &
IoD, Autonomous UAV Missions &
LLM integrated with RAG over a knowledge graph built from real-time sensor data and historical mission logs; fully auditable decision tracing &
92\% decision accuracy; 94\% mission success rate in dynamic environments \\
\hline
\cite{sezgin2025scenario} &
Augmented LLM--RAG IoD Decision Model &
IoD, Multi-UAV Network Coordination &
Centralized LLM with RAG processing telemetry, mission logs, and environmental data for real-time contextual decisions &
BLEU score: 0.82; cosine similarity: 0.87; average decision latency: 120 ms \\
\hline
\cite{cai2025llm} &
LLM-Land &
Autonomous UAV Landing &
Lightweight LLM with RAG combined with a Vision-Language Encoder (VLE) and Model Predictive Control (MPC) for semantic constraint-aware trajectory replanning &
High landing success rate; superior obstacle avoidance compared to baseline MPC and non-RAG LLM-MPC; real-time feasibility on constrained hardware \\
\hline
\cite{afzal2025intelligent} &
LLM--RAG UAV-BS Optimization Framework &
UAV-assisted Mobile Networks (UAV-BSs) &
GAI-assisted LLM with RAG (e.g., GPT-4) to automate mathematical modeling and optimize UAV placement strategies &
Faster convergence and higher efficiency than Genetic Algorithm-based approaches \\
\hline
\cite{peng2024graph} &
GraphRAG &
General RAG Enhancement (incl. UAV Analytics) &
Knowledge-graph-based retrieval of nodes, subgraphs, and communities instead of isolated text chunks &
Improved query-focused summarization; reduced redundancy; enhanced global domain understanding \\
\hline
\cite{sarmah2024hybridrag} &
HybridRAG &
Domain-Specific Analytics (e.g., UAV Systems) &
Combination of VectorRAG for semantic similarity and GraphRAG for structured relational reasoning &
More faithful, comprehensive, and domain-aligned LLM responses \\
\hline
\cite{wen2025hybridrag} &
Advanced HybridRAG for UAV MEC &
Multi-UAV Mobile Edge Computing, Carbon Emission Optimization &
Integrated KeywordRAG, VectorRAG, and GraphRAG, enabling hybrid factual and relational reasoning &
Significantly improved reliability and precision in complex optimization model formulation \\
\hline
\cite{11194943} &
Enterprise-Grade Modular RAG--LLM Architecture &
Intelligent Robotic Control &
Modular RAG pipeline using ChromaDB and Docling with LLMs (e.g., Claude~3.5), grounded in manuals, logs, and safety validators &
Improved instruction grounding and operational adaptability; challenges in cloud dependence and computational cost \\
\hline
\cite{es2024ragas} &
RAGAS &
RAG Evaluation &
Reference-free automated evaluation of faithfulness, answer relevance, and context relevance &
High agreement with human judgment, particularly for faithfulness \\
\hline
\cite{ru2024ragchecker} &
RAGCHECKER &
RAG Evaluation and Diagnostics &
Fine-grained, component-level evaluation disentangling retrieval and generation quality &
Stronger correlation with human judgments; effective identification of retrieval--generation trade-offs \\
\hline
\end{tabular}%
}
\end{table*}


\begin{figure}[!t]
  \centering
  \includegraphics[width=0.35\textwidth]{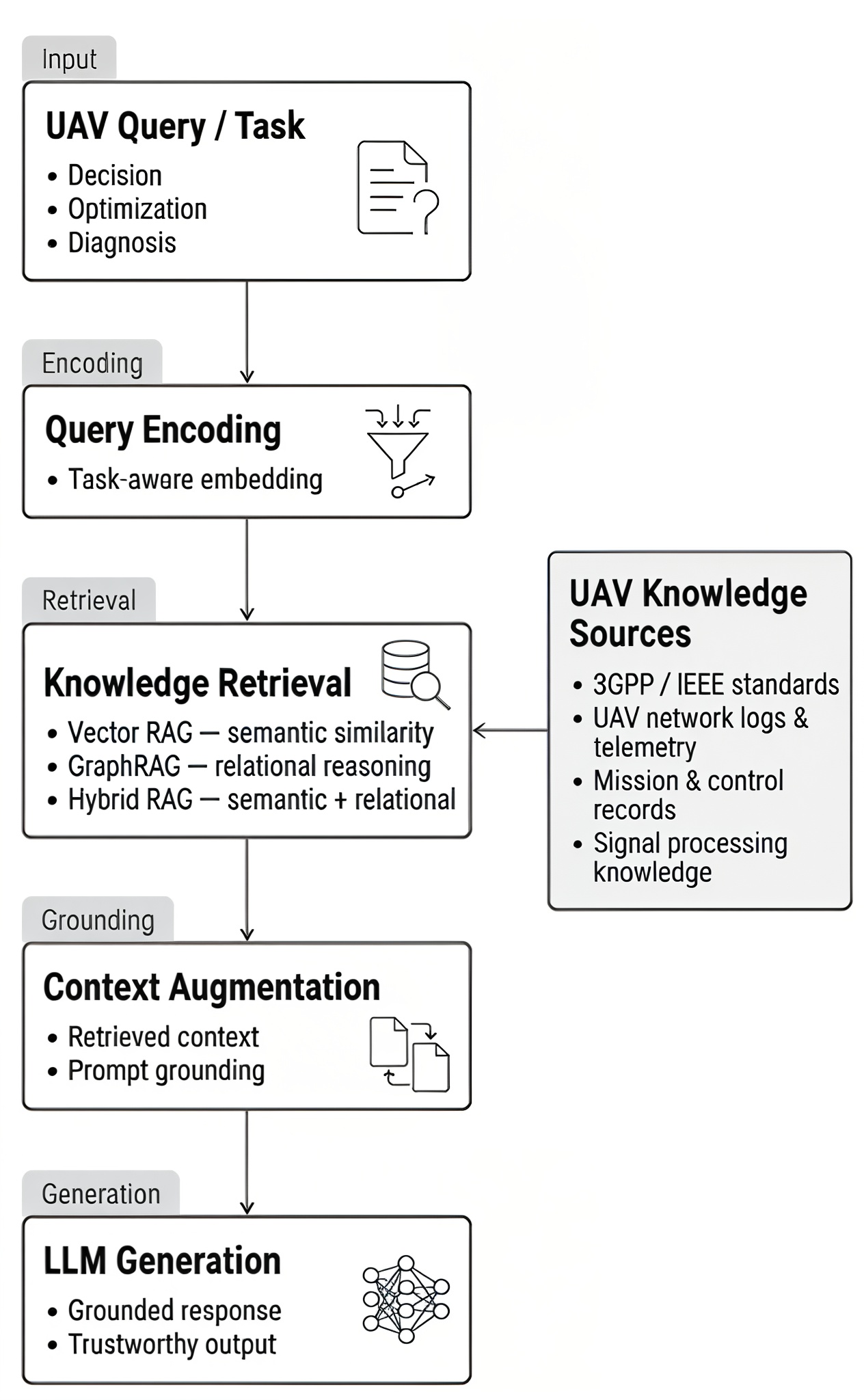} 
  \caption{Conventional RAG pipeline for UAV applications, illustrating the workflow from UAV task/query formulation through task-aware query encoding, hybrid knowledge retrieval (vector, graph, and hybrid RAG), context augmentation, and grounded LLM-assisted response generation using domain-specific UAV knowledge sources.
}
  \label{fig:rag}
\end{figure}

\subsection{Retrieval-Augmented Generation for UAVs}

RAG offers an effective approach to adapting LLMs to the dynamic and rapidly evolving field of UAV networks by grounding generation in external, up-to-date knowledge. Unlike conventional LLMs that depend solely on static training data, RAG retrieves relevant information at inference time, allowing models to incorporate evolving standards, architectures, and transmission technologies critical to UAV networking. Access to authoritative sources such as 3GPP and IEEE standards, emerging network paradigms, and recent advances in signal processing significantly improve the factual accuracy and contextual relevance of the generated outputs.
A conventional RAG pipeline, as shown in Fig.~\ref{fig:rag}, typically includes several stages: document indexing in vector databases, query processing and embedding, retrieval and ranking of relevant documents, prompt augmentation with retrieved context, and final response generation. These components together balance retrieval accuracy, response latency, and system scalability.

RAG improves LLM performance in UAV networks by grounding responses in up-to-date, domain-aligned external knowledge. As a result, it offers a cost-effective and flexible alternative to extensive fine-tuning, enabling reliable and adaptive intelligence in rapidly evolving UAV communication and networking environments.
For example, Sezgin \textit{et al.} \cite{sezgin2025llm, sezgin2025scenario} investigate LLM--RAG–based decision-making frameworks for the Internet of Drones (IoD), demonstrating how grounding LLM reasoning in real-time sensor data and structured mission logs enables context-aware, interpretable, and timely UAV control. Their studies report strong performance in dynamic environments, including up to 92\% decision accuracy, 94\% mission success rate, and sub-second latency (e.g., 120 ms), highlighting the feasibility of RAG-enhanced LLMs for real-time IoD applications.
Beyond high-level decision-making, Cai \textit{et al.} \cite{cai2025llm} propose the LLM-Land, a hybrid framework combining RAG-enhanced LLMs with vision--language encoding and model predictive control for autonomous UAV landing in cluttered environments. By translating visual observations into semantic constraints for control, the framework outperforms conventional MPC-assisted baselines while remaining deployable on resource-constrained hardware. Similarly, Afzal \textit{et al.} \cite{afzal2025intelligent} applied RAG-enhanced LLMs to UAV-assisted mobile networks, showing that natural-language-driven optimization can accelerate convergence and improve efficiency compared with traditional heuristic methods such as genetic algorithms.
\par
Despite these successes, conventional RAG suffers from inherent limitations, including fragmented retrieval, redundancy, and limited modeling of inter-document relationships. To address these issues, GraphRAG \cite{peng2024graph} retrieves structured graph elements rather than isolated text chunks, capturing relational dependencies and improving performance on complex reasoning tasks. Building on this idea, HybridRAG combines vector-based semantic retrieval with graph-based relational retrieval to provide richer grounding for domain-specific applications \cite{sarmah2024hybridrag}. Wen \textit{et al.} \cite{wen2025hybridrag} further extend HybridRAG by integrating KeywordRAG, VectorRAG, and GraphRAG to support multi-UAV mobile edge computing, thereby enabling a more reliable formulation of complex carbon-emission optimization problems.

Collectively, these works illustrate a clear evolution: from standard RAG-based UAV optimization, to graph-structured retrieval, to hybrid retrieval systems, and ultimately to fully integrated, enterprise-grade RAG–LLM architectures for autonomous and robotic control. However, evaluating RAG systems remains a significant challenge due to their modular design, the complexity of long-form outputs, and the difficulty of obtaining reliable ground truth. To address this, Es \textit{et al.} \cite{es2024ragas} introduced the Retrieval Augmented Generation Assessment (RAGAS). This reference-free evaluation framework measures three core dimensions: faithfulness, answer, and context relevance by focusing on factual grounding, query alignment, and retrieval quality. RAGAS enables rapid, automated evaluation with high agreement with human judgments, particularly in terms of faithfulness. Table \ref{tab:rag_uav_summary} provides a summary of RAG frameworks for UAV operation and communications.

\begin{figure*}[t]
  \centering
  \includegraphics[width=0.9\textwidth]{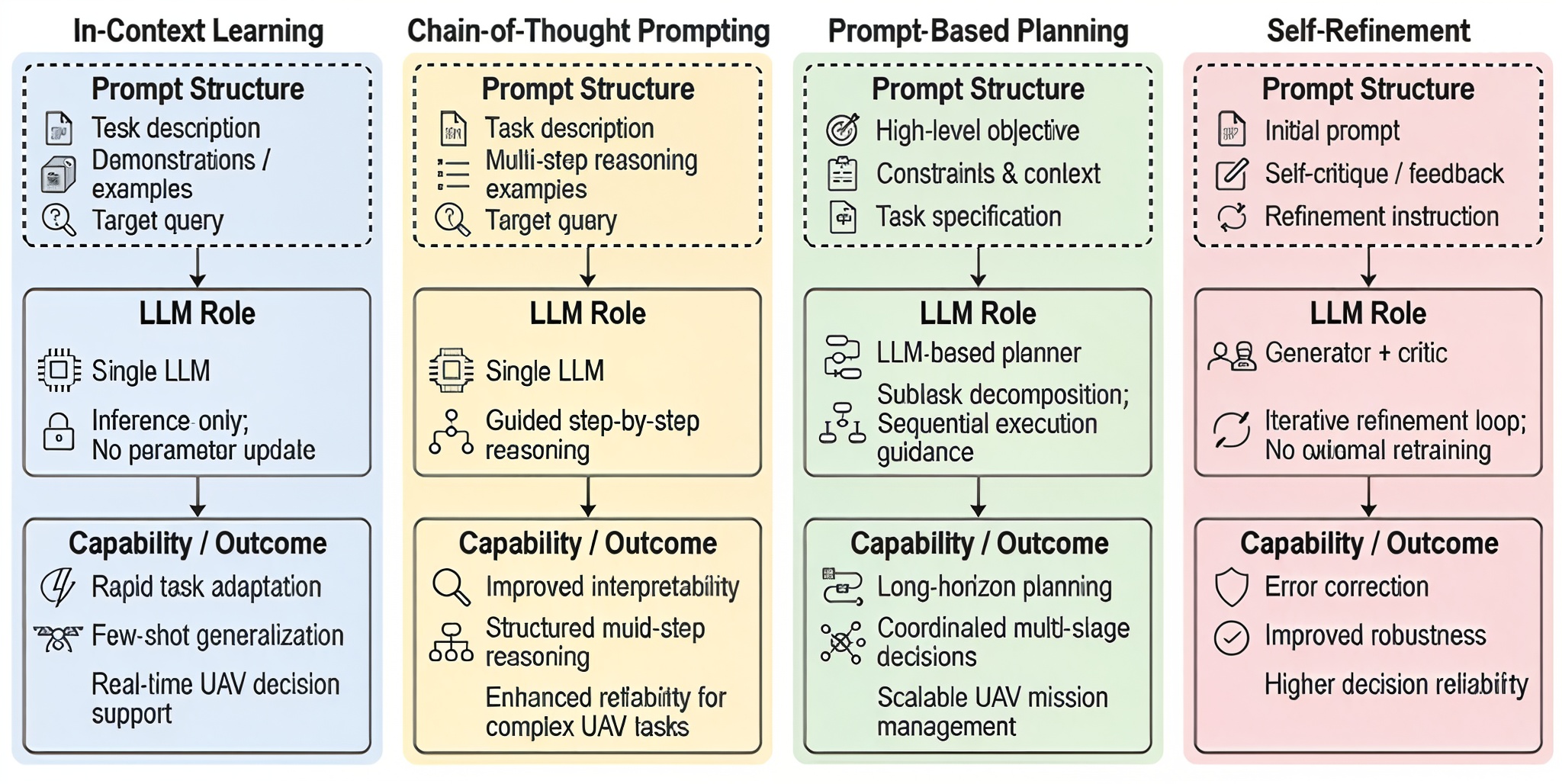} 
  \caption{Prompt engineering strategies for UAV applications, comparing in-context learning, chain-of-thought prompting, prompt-based planning, and self-refinement in terms of prompt structure, LLM roles, and resulting capabilities for reliable, scalable, and robust UAV decision-making.
}
  \label{fig:prompt}
\end{figure*}

\subsection{Prompt Engineering: Strategies for UAV Intelligence}

Prompt engineering is the process of designing prompts to guide LLMs in generating accurate and useful outputs. As a reusable and adaptable methodology, prompt engineering supports structured problem-solving across diverse domains, improves understanding of LLM behavior, and enables few-shot learning by optimizing in-context prompts. It also plays an important role in the development of advanced UAV systems. As LLM technologies evolve, prompt engineering is expected to become increasingly important, offering finer control over model outputs and enabling new classes of applications. The core techniques of prompt engineering, including ICL, CoT prompting, prompt-based planning, and self-refinement, are depicted in Fig. \ref{fig:prompt} and are discussed below.

\subsubsection{In-Context Learning - ICL}

ICL enables LLMs to adapt to target tasks by conditioning on natural-language instructions and demonstrations provided directly in the prompt, without modifying model parameters. In UASNets, ICL operates by providing contextualized demonstrations, such as prior network states, scheduling decisions, and environmental conditions, that allow the model to exploit pretrained knowledge for real-time adaptation. The effectiveness of ICL depends heavily on the selection and formatting of demonstrations, requiring domain expertise in both UAV networking and LLM behavior.

In dynamic UAV networking and emergency scenarios, ICL enables LLMs to incorporate real-time sensory feedback—such as queue lengths, channel conditions, battery levels, AoI, and UAV locations—directly into prompts for adaptive decision-making \cite{emami2025llm}.
A series of works by Emami \textit{et al.} demonstrate the effectiveness of this paradigm across multiple UAV applications. ICLDC applies ICL to emergency data-collection scheduling in UASNets, reducing packet loss and computational overhead compared with DQN and heuristic baselines \cite{emami2025llm}. FRSICL extends this idea to joint flight-resource allocation for wildfire monitoring, achieving lower AoI and improved stability compared to PPO-based methods \cite{emami2025frsicl}. Further generalization to broader disaster-response tasks—including path planning, scheduling, velocity control, and power management—highlights both performance gains and potential prompt-security vulnerabilities \cite{emami2025prompts}. In multi-UAV coordination, the attention-assisted AIC-VDS framework compresses high-dimensional sensory inputs to support joint scheduling and velocity optimization, significantly outperforming multi-agent DRL baselines \cite{emami2025joint}.
Beyond networking optimization, Ji \textit{et al.} integrate ICL with human-in-the-loop verification in the LLM-CRF framework for human–swarm teaming, enabling hierarchical task decomposition and coordinated control in SAR missions while improving mission success rates and reducing operator workload \cite{ji2025llm}.

Collectively, these studies indicate that ICL provides a lightweight, adaptive alternative to conventional learning-based optimization, enabling real-time, interpretable, and scalable decision-making across UAV networking, resource management, and swarm coordination tasks.

\begin{table*}[t]
\centering
\caption{ICL frameworks leveraging LLMs for emergency UAV and sensor network operations, comparing approaches for data collection, wildfire monitoring, disaster response, multi-UAV coordination, and human--swarm teaming in search and rescue.}
\label{tab:icl_emergency_uav}
\renewcommand{\arraystretch}{1.25}
\resizebox{\textwidth}{!}{%
\begin{tabular}{|
m{1cm}|
m{1.5cm}|
m{2.6cm}|
m{4.8cm}|
m{4.8cm}|
m{4.8cm}|}
\hline
\textbf{Reference} &
\textbf{Proposed Framework} &
\textbf{Primary Application Domain} &
\textbf{Core Function of the LLM} &
\textbf{Key Advantage over Traditional Methods} &
\textbf{Key Performance Results} \\
\hline
\cite{emami2025llm} &
ICLDC &
UASNets, SAR Emergencies &
Edge-deployed LLM uses ICL to generate data collection schedules from natural-language task descriptions derived from sensory logs &
Eliminates iterative DRL training; enables rapid adaptation, improved safety via verifier, and robustness through attack detection &
Significant reduction in cumulative packet loss vs. DQN and Maximum Channel Gain; faster convergence and lower computational overhead \\
\hline
\cite{emami2025frsicl} &
FRSICL &
UAV-Assisted Wildfire Monitoring &
LLM performs real-time flight resource allocation by jointly optimizing data collection schedules and UAV velocities using contextual task descriptions &
Joint optimization without retraining; improved responsiveness and stability compared to policy-gradient and heuristic methods &
Lower average AoI and more stable velocity control than PPO and Nearest-Neighbor baselines \\
\hline
\cite{emami2025prompts} &
LLM-Assisted ICL for Public Safety UAVs &
Disaster Response and Wildfire Monitoring &
Edge LLM leverages prompts and demonstrations to adaptively control UAV path planning, velocity, scheduling, and power management &
Generalizes across tasks without retraining; flexible prompt-based control versus task-specific optimization models &
Significant packet loss reduction in data collection; highlights resilience and exposure to jailbreaking vulnerabilities \\
\hline
\cite{emami2025joint} &
AIC-VDS &
Multi-UAV Post-Disaster Monitoring (e.g., Tsunami Response) &
Attention-guided LLM focuses on urgent sensors to generate task descriptions for joint scheduling and velocity control &
Efficient compression of high-dimensional sensory data; scalable coordination beyond multi-agent DRL limits &
Up to 91\% packet loss reduction compared to Multi-agent DQN and Maximum Channel Gain baselines \\
\hline
\cite{ji2025llm} &
LLM-CRF &
Human--Swarm Teaming for Disaster SAR &
Central LLM interprets multi-modal human commands, extracts intent, and performs hierarchical task decomposition and mission planning &
Transforms human role from low-level control to high-level supervision; improves safety with ICL and HITL verification &
64.2\% reduction in task completion time; 94\% mission success rate; 42.9\% reduction in operator cognitive workload \\
\hline
\end{tabular}%
}
\end{table*}

\subsubsection{Chain-of-Thought Reasoning}

CoT reasoning represents a significant advance in AI systems' ability to handle complex reasoning tasks by explicitly decomposing problems into intermediate steps \cite{wei2022chain}. Instead of generating answers directly, CoT prompts models to reason step-by-step, mirroring human problem-solving processes and improving performance on tasks involving mathematics, logic, planning, and multi-step decision-making. By revealing intermediate reasoning, CoT enhances interpretability and explainability, improves generalization, and helps mitigate hallucinations by constraining generation to verifiable logical sequences.

Since its introduction, CoT prompting has evolved from basic paradigms such as Zero-shot CoT and Few-shot CoT to more advanced and scalable frameworks. These include Auto-CoT, which automatically selects effective demonstrations, and Self-Consistency CoT, which aggregates multiple reasoning paths to improve reliability. More recent extensions, such as Tree-of-Thought (ToT) and Graph-of-Thought (GoT), further structure reasoning as trees or graphs, enabling exploration and comparison of alternative solution trajectories. Although these advanced methods introduce additional computational overhead, they are particularly valuable for complex, intent-driven tasks, such as UAV networking,  where transparent, logical, and verifiable decision-making is essential \cite{wang2025chain}.
Zhu \textit{et al.} \cite{zhu2025novel} demonstrate the effectiveness of CoT in UAV swarm decision-making by integrating an LLM with an RL controller. In their framework, CoT prompting enables the LLM to decompose high-level missions into structured sub-tasks and dynamically adjust the RL reward function based on real-time environmental feedback.

Overall, these studies indicate that CoT reasoning enhances adaptability, robustness, and decision-making accuracy in UAV systems operating under uncertainty and rapid dynamics. The progression from simple prompting to structured reasoning architectures positions CoT as a powerful tool for complex UAV tasks where logical transparency and verifiability are critical, even at the cost of increased computational complexity.

\subsubsection{Prompt-Based Planning}

For large-scale, complex tasks such as network configuration or project development, prompt-based planning breaks down the overall objective into a sequence of manageable sub-tasks. In this approach, the LLM first generates a structured plan of ordered sub-tasks, which are then executed sequentially. This enables the model to address problems that require multi-step reasoning and the integration of information from multiple sources. Prompt-based planning further improves problem-solving by guiding the LLM through explicit action sequences, incorporating intermediate feedback, and continuously refining the plan.

This capability is especially important for UAV operations, where tasks such as formation reconfiguration, resource allocation in multi-UAV systems, and coordinated mission planning require long-horizon reasoning across multiple agents in dynamic environments. Therefore, designing effective prompts and planning structures is essential to the successful deployment of LLMs in real-time UAV control and decision-making applications.
Although prompt-based planning enables LLMs to decompose complex objectives into executable subtasks, the initial plans and decisions produced by the model may still be imperfect or suboptimal. Therefore, self-refinement is essential.

\begin{table*}[t!]
\centering
\caption{Comparison of LLM adaptation techniques—pre-training, fine-tuning, prompt engineering, and RAG—in terms of their core objectives, inputs, processes, resource requirements, advantages, challenges, suitability for UAV applications, and basis of output generation.}
\renewcommand{\arraystretch}{1.25}
\label{tab:llm_adaptation_comparison}
\begin{tabular}{|
m{2.2cm}|
m{3cm}|
m{3cm}|
m{4cm}|
m{4cm}|}
\hline
\textbf{Aspect} &
\textbf{LLM Pre-training} &
\textbf{LLM Fine-tuning} &
\textbf{Prompt Engineering} &
\textbf{RAG} \\
\hline

\textbf{Core Objective} &
Learn general language patterns, knowledge, and reasoning abilities from a massive corpus. &
Adapt a pretrained model's knowledge to a specific task or domain using a smaller dataset. &
Guide the output of a pretrained model for a specific task by carefully designing the input prompt. &
Enhance the output of a pretrained model by retrieving and incorporating relevant, up-to-date information from an external knowledge base. \\
\hline

\textbf{Primary Input} &
Massive, general/specialized text corpora (e.g., web data, books, code). &
Smaller, task-specific, often labeled datasets. &
A carefully crafted textual prompt (instructions, examples, context). &
User query + relevant information retrieved from an external knowledge base. \\
\hline

\textbf{Core Process} &
Unsupervised next-token prediction on a vast dataset. Training from scratch. &
Updating model parameters (fully or via PEFT) on a targeted dataset. &
Designing and iteratively refining the input text. No weight updates. &
Retrieving relevant documents and augmenting the prompt with this context before generation. \\
\hline

\textbf{Resource Intensity} &
Extremely High. Vast data, immense compute (GPU clusters, 3D parallelism). &
Moderate to High. Lower than pre-training but still significant for full fine-tuning. &
Very Low. Requires only a forward pass; no training. &
Low to Moderate. Adds a retrieval step and a longer context, but inference-only. \\
\hline

\textbf{Key Advantages} &
Develops foundational language capabilities and emergent skills. &
High task-specific performance, efficient transfer learning. &
Highly flexible, fast to implement, resource-efficient. &
Provides up-to-date \& factual information, reduces hallucinations. \\
\hline

\textbf{Key Challenges} &
Prohibitively expensive/time-consuming. Requires vast domain-specific data. &
Computationally heavy for edge systems. Requires quality labeled data. &
Effectiveness depends heavily on the design of prompts. &
Requires building/maintaining a high-quality, updated knowledge base. \\
\hline

\textbf{UAV Application Suitability} &
Foundational models via large-scale collaboration (e.g., Federated Learning). &
Specialized models for mission planning or diagnosis with sufficient data. &
Ideal for real-time, low-latency tasks (path planning, collision avoidance). &
Highly promising for dynamic networking tasks with the latest protocols/data. \\
\hline

\textbf{Output Basis} &
Model's internal parameters from initial training. &
pretrained parameters adjusted by fine-tuning data. &
The model's inherent capabilities are triggered by the prompt. &
Synthesis of the model's knowledge and retrieved external information. \\
\hline
\end{tabular}
\end{table*}

\subsubsection{Self-Refinement} 

Self-refinement is an emerging paradigm for improving the initial outputs through an iterative process of self-critique and refinement. In this approach, an LLM evaluates its own outputs and incrementally improves them based on internally generated feedback. For example, if an initial decision, such as a data collection schedule, is suboptimal or incorrect, the same model can be prompted to analyze its own output, identify deficiencies, and generate a revised solution. This significantly reduces the need for human intervention, which is particularly valuable given the complexity of networking and UAV-related tasks \cite{zhou2025large}.
\par
A salient merit of self-refinement is that it requires only a single LLM, which simultaneously serves as the generator, critic, and refiner, without relying on additional training data or RL. The process typically begins with the generation of an initial output, followed by a self-critique stage in which the model provides feedback on its own reasoning or results. This feedback is then used to generate an improved output, and the cycle is repeated until a predefined stopping criterion, such as convergence, task completion, or iteration limits, is met \cite{madaan2023self}.
\par
Building on this concept, Wang \textit{et al.} \cite{wang2025enhanced} propose a closed-loop framework that significantly enhances the reliability and reasoning capabilities of LLMs for autonomous UAV control. Their approach combines a structured prompt framework, Guidelines, Skill APIs, Constraints, and Examples (GSCE), with an iterative feedback and refinement mechanism. After the LLM generates an initial control program using the GSCE framework, the code is executed within a simulator to produce a UAV trajectory. A dedicated LLM-assisted discriminator then evaluates the trajectory against the task objectives and generates targeted diagnostic feedback if deviations are detected. This feedback is subsequently used to guide the GSCE-guided LLM, prompting code refinement and regeneration. The loop of generation, execution, evaluation, and refinement continues until the task is completed or the predefined iteration limit is reached. Experimental results demonstrate that this closed-loop self-refinement system substantially outperforms one-shot generation approaches, achieving task success rates exceeding 90\% by enabling the LLM to learn from execution errors and progressively correct its outputs.

\subsection{Key Findings and Insights}

Adapting LLMs to UAV systems requires carefully balancing performance, computational efficiency, and domain specificity. 
The following provides a comprehensive discussion of these techniques, examining their trade-offs, recent advancements, and future research directions for scaling LLMs to meet the stringent operational demands of UAVs in mission-critical and highly dynamic environments.

\subsubsection{\textbf{Edge, On-Device, and Hybrid LLM Deployment}}

LLMs offer strong reasoning and generalization capabilities that can enhance UAV autonomy in navigation, surveillance, and mission-critical tasks such as SAR and ISR. However, most existing deployments rely on cloud infrastructure, which introduces latency, bandwidth overhead, and privacy concerns, making purely cloud-centric solutions unsuitable for delay-sensitive or communication-constrained UAV operations.

\begin{enumerate}[label=\alph*)]

\item \textbf{Centralized Edge Inference:} UAVs offload sensor data to an edge server hosting the LLM. Although communication-intensive, techniques such as model compression and token reduction can mitigate overhead. With proper optimization (e.g., quantization, pruning, parameter-efficient tuning), edge inference can support near–real-time decision-making for safety-critical tasks.

\item \textbf{Local Inference (On-Device):} Running LLMs directly on UAV hardware improves privacy, reduces latency, and enhances robustness to connectivity loss. Lightweight deployments, such as Aero-LLM \cite{dharmalingam2025aero}, demonstrate the feasibility of using small models (e.g., OPT-125M/350M). However, onboard inference is constrained by limited computational resources, high energy consumption, and limited memory capacity, which often limit reasoning depth for complex tasks such as trajectory planning and resource allocation \cite{lin2025pushing}.

\item \textbf{Split Inference (Model Partitioning):} The model is divided across UAV and edge/cloud platforms. Early layers are executed onboard to process sensor inputs, while deeper layers are executed remotely. This reduces onboard computation and transmits compact intermediate representations.

\item \textbf{Collaborative Device–Server Inference:} Lightweight onboard models cooperate with powerful remote LLMs. Methods such as speculative decoding allow local draft generation followed by server-side verification, reducing bandwidth compared with full data offloading and improving robustness under limited connectivity \cite{ping2025multimodal}.

\end{enumerate}

Each paradigm entails trade-offs among latency, energy consumption, memory constraints, and communication overhead. The practicality of LLM deployment in UAV systems ultimately depends on hardware capability, network conditions, and mission requirements. 

\subsubsection{\textbf{Balancing Model Size, Memory, and Performance on the Edge}}

The scalability of LLMs on edge hardware is fundamentally constrained by memory capacity, computational throughput, and energy availability, all of which are significantly more limited than in cloud environments. A theoretical upper bound on the number of model parameters (P) that can be accommodated on an edge device can be approximated as
$P = \frac{M \times 8}{b}$, where $M$ denotes the available memory in bytes and $b$ represents the bit precision per parameter. 
In practice, however, this upper bound is substantially reduced by runtime overheads, activation storage, and intermediate buffers, thereby reducing the required capacity to roughly 1–10 billion parameters. This practical range is consistent with the scale of current state-of-the-art on-device models, such as Gemini Nano \cite{GoogleGemini2025} and Qualcomm’s LLMs with up to 10 billion parameters \cite{xiao2024large}. Consequently, most real-world edge deployments operate in the 1–3 billion-parameter regime, whereas larger models require cloud offloading or distributed execution across multiple devices \cite{liu2025survey, giorgetti2025transitioning}.
Given the stringent memory, computational, and energy constraints of edge platforms, achieving real-time LLM inference on UAVs typically requires not only careful model scaling and optimization, but also high-speed, low-latency communication with edge or cloud servers.

\subsubsection{\textbf{Structuring Knowledge for Reliable Retrieval}}

Chunking is a fundamental component of modern RAG systems. As LLM context windows continue to grow, the primary bottleneck has shifted from generation to retrieval, making the structure and granularity of chunks central to overall system performance. Effective chunking breaks information into coherent and meaningful units that reflect how humans naturally organize and understand ideas, rather than relying on arbitrary text boundaries. This distinction is critical because the retrieval quality determines the context the model ultimately sees. Most RAG hallucinations arise from missing, fragmented, or poorly segmented information.
Different chunking strategies, such as fixed-length, hierarchical, semantic, sliding-window, sentence-based, topic-boundary, and hybrid approaches, offer distinct strengths and tradeoffs, and their effectiveness depends heavily on the structure and nature of the underlying content. Ultimately, chunking is not merely a preprocessing step; it functions as an information architecture for AI, shaping how models perceive, retrieve, and reason over knowledge. Better chunking leads directly to more accurate, efficient, and reliable RAG systems.

\subsubsection{\textbf{Quantization as an Inference Optimization Technique for Efficient LLMs}}

Quantization is a widely used model compression technique that improves the efficiency of LLMs by reducing the numerical precision of their parameters, and, in some cases, their activations, from high-precision floating-point formats (e.g., FP32 or FP16) to lower-bit representations such as INT8, INT4, or INT3. This reduction in precision substantially decreases model size, memory footprint, computational complexity, inference latency, and energy consumption, making quantization particularly important for deploying LLMs on resource-constrained edge devices, including Raspberry Pi–class hardware. Quantization can be applied after training via Post-Training Quantization (PTQ), which converts the pretrained model weights without retraining, or during training via Quantization-Aware Training (QAT), which simulates low-precision arithmetic to better preserve accuracy. Despite these efficiency gains, quantization introduces trade-offs: aggressive reductions in bit width may lead to accuracy degradation, numerical instability, or additional dequantization overhead, particularly for tasks involving complex reasoning or arithmetic. As a result, effective quantization requires carefully balancing efficiency gains with acceptable performance loss to enable practical, sustainable edge AI deployment.

In summary, building on the previous discussion of LLM training, adaptation, and architectural choices, it becomes clear that these models provide a flexible and scalable foundation for UAV operations. Due to their high mobility and agile maneuvering, UAVs have been widely adopted for civil and commercial applications, including environmental monitoring, public safety, and parcel delivery. Beyond these roles, UAVs increasingly function as airborne data relays or aerial base stations in infrastructure-limited or disaster-stricken environments, dynamically establishing reliable LoS links with ground sensors to enhance throughput, extend coverage, and reduce operational risk. LLMs can leverage this inherent mobility and the heterogeneous, real-time nature of aerial networks to support communication-aware decision-making, trajectory optimization, swarm coordination, and safety management—tasks that conventional optimization or RL methods often handle inefficiently. 

\section{LLM for UAV Communications} \label{sec4}

Owing to their high mobility and flexible deployment, UAVs have been widely adopted in civil and commercial applications, including environmental monitoring, traffic management, parcel delivery, and precision agriculture~\cite{ma2018experiment,rumba2020wild,huang2019optimal,lin2020energy}. This section examines how LLMs enhance UAV communications across four key dimensions. It first explores LLM-assisted navigation, trajectory planning, and placement strategies for communication-aware UAV deployment. It then discusses swarm-level control and coordinated execution enabled by language-driven reasoning. Next, it analyzes how LLMs contribute to UAV safety, autonomy, and network optimization under dynamic and resource-constrained conditions. Finally, it reviews multi-LLM architectures designed to support scalable, collaborative, and communication-efficient UAV systems.

\begin{figure*}[h]
  \centering
  \includegraphics[width=\textwidth]{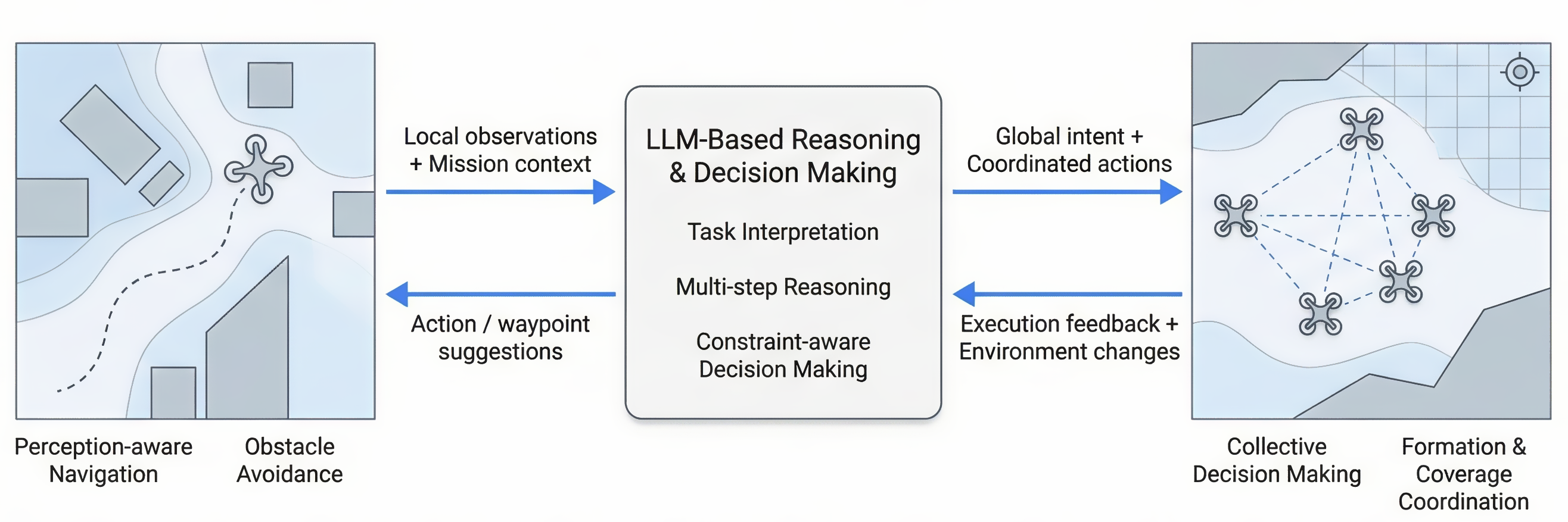} 
  \caption{Diagram of the LLM-assisted UAV navigation, planning, and placement framework, illustrating the integration of mission context, reasoning, decision-making, and multi-agent coordination for autonomous operations.}
  \label{fig:uav_navi}
\end{figure*}

\subsection{LLM-Assisted UAV Navigation, Planning, and Placement}

UAV navigation and mission-level planning are fundamental to autonomous SAR and ISR operations, as they determine where UAVs should move, how coverage should be allocated, and how multiple aerial assets should be positioned to accomplish mission objectives. In complex and dynamic environments, these planning decisions must account for environmental constraints, mission priorities, communication limitations, and regulatory requirements, often under strict time constraints.
\par
Traditional approaches to UAV navigation and multi-UAV planning, such as combinatorial optimization and RL, typically involve high computational costs and long convergence times, limiting their practicality for real-time or large-scale deployment. Other learning-based methods, such as Deep Neural Networks (DNNs) and heuristic algorithms, often lack interpretability and struggle to adapt to rapidly changing mission contexts. Recent advances in LLMs provide a new paradigm for addressing these challenges by enabling high-level reasoning, semantic understanding, and iterative refinement of planning decisions using natural-language task descriptions, historical solutions, and performance feedback. Fig. \ref{fig:uav_navi} depicts LLM-Assisted UAV Navigation, Planning, and Placement.

\par
In this subsection, we focus on the use of LLMs at the planning and decision-making layer, including UAV navigation, swarm-level path planning, and multi-UAV placement optimization. Rather than addressing low-level control or execution, the approaches discussed use LLMs to generate, evaluate, and refine high-level plans that guide downstream motion-planning and control modules. This perspective highlights the role of LLMs as cognitive planners that complement traditional optimization and learning techniques in complex UAV missions.
\par
Clader \textit{et al.}\cite{cladera2025air} present a novel air–ground robotic system in which a UAV and an Unmanned Ground Vehicle (UGV) collaborate to execute missions specified in natural language within unknown, kilometer-scale environments. Their system employs an LLM-assisted planner, SPINE, to interpret high-level user intent, infer relevant semantic objects, and dynamically generate mission plans. A salient innovation is the construction of a compact, open-set semantic–metric map, built online by the UAV and opportunistically shared with the UGV via the MOCHA communication framework. This design enables effective coordination despite intermittent connectivity. Experimental results in both urban and rural environments demonstrate the system’s ability to complete seven distinct language-specified missions, such as inspecting a blocked bridge or triaging a chemical spill, highlighting its robustness in large-scale, unstructured settings. While this work focuses on air–ground cooperation for single-mission execution, subsequent studies have explored how LLM-assisted reasoning can scale to more complex multi-agent aerial systems.
\par
In this direction, Xiao \textit{et al.}\cite{xiao2025llm} propose an LLM-assisted framework for UAV swarm path planning in Industry 5.0 scenarios, such as low-altitude rescue missions. The approach addresses the limitations of traditional DNNs and heuristic algorithms in terms of interpretability and adaptability to dynamic environments. The core of the method is the Multi-step Deep Thinking Movement Decision Samples Generation (MSDTMD-SG) algorithm, which creates fine-tuning datasets by simulating human-like multi-step reasoning. Using symbolic modeling and a DRL-inspired reward function, the algorithm identifies optimal movement positions from local observations and embeds the reasoning process into structured templates for LLM training. A cloud–edge collaboration framework enables UAVs to use a cloud-based large LLM when communication is stable and a compressed onboard small LLM when connectivity is limited. Additionally, a scene memory and replay mechanism supports continual learning from failures. Experimental evaluations show that the proposed method achieves the highest task completion rate compared to DRL and Ant Colony Optimization (ACO) baselines, while maintaining competitive computational cost and system latency, and providing superior adaptability in dynamic industrial environments.
\par
Beyond trajectory and path planning, researchers have explored LLM-assisted approaches for high-level multi-UAV decision-making, focusing on placement and coordination rather than low-level swarm control.
Wang \textit{et al.}\cite{wang2025multi} introduce an LLM-assisted framework for optimizing the placement of multiple UAVs to establish Integrated Access and Backhaul (IAB) networks. Instead of relying on computationally intensive combinatorial optimization or RL methods, the proposed approach employs iterative structured prompting, enabling the LLM to generate and refine candidate UAV positions using problem descriptions, historical solutions, and performance feedback. Simulation results demonstrate that the method achieves near-optimal connectivity and coverage with significantly fewer iterations than conventional techniques, making it suitable for rapid deployment in dynamic environments.
\par
Collectively, these studies demonstrate the versatility of LLMs in enhancing UAV capabilities across a broad spectrum, from single-robot air–ground coordination to swarm-level reasoning and large-scale placement optimization. However, most existing work focuses on isolated components of the autonomy pipeline rather than providing a unified end-to-end solution. This limitation motivates broader efforts to integrate natural-language task interpretation with the full execution stack.
\par
Addressing this gap, Yuan \textit{et al.}\cite{yuan2025next} introduce Next-Generation LLM for UAV (NELV), the first end-to-end framework that directly links natural-language instructions to fully executable UAV flight trajectories across short-, medium-, and long-range missions. NELV integrates five key components, LLM-as-Parser, route planning, path planning, control execution, and real-time monitoring, to enable complex tasks such as multi-UAV patrol, multi-point-of-interest delivery, and long-distance multi-hop relocation. Through three representative use cases, the framework demonstrates how LLMs can significantly reduce pilot workload while adhering to regulatory, airspace, and environmental constraints. Furthermore, the authors propose a five-level autonomy roadmap that outlines the progression from current language-based instruction parsing to fully autonomous LLM-as-Autopilot systems, identifying critical challenges related to regulation-aware reasoning, data scarcity, and safety-critical evaluation.
\par
Recent frameworks such as NELV demonstrate a promising path toward fully autonomous and intelligent UAV operations by connecting natural-language tasking with end-to-end flight execution, including route planning, path generation, control, and real-time monitoring. However, extending these capabilities to large-scale UAV swarms introduces additional challenges for coordination and control. Using LLMs for intelligent, scalable, and context-aware swarm coordination offers a compelling approach to addressing these challenges and advancing the next generation of autonomous UAV systems.

\begin{table*}[t]
\centering
\caption{Comparative overview of recent LLM-assisted UAV planning, coordination, and swarm control frameworks, highlighting their autonomy layers, application domains, core LLM functions, and demonstrated performance benefits across diverse mission scenarios.}
\label{tab:llm_uav_planning_control}
\resizebox{\textwidth}{!}{%
\begin{tabular}{|
m{0.5cm}|
m{1.8cm}|
m{2cm}|
m{2.6cm}|
m{4cm}|
m{4cm}|
}
\hline
\textbf{Ref.} &
\textbf{Proposed Framework} &
\textbf{Autonomy Layer} &
\textbf{Primary Application Domain} &
\textbf{Core Function of the LLM} &
\textbf{Key Results / Advantages} \\
\hline

\cite{cladera2025air} &
SPINE-based Air--Ground System &
Mission-Level Planning &
Air--Ground Cooperative Robotics in Unknown Environments &
Interprets natural language missions, infers semantic objects, and generates adaptive plans using online semantic--metric mapping &
Successfully completes language-specified missions in kilometer-scale urban and rural environments under intermittent connectivity \\
\hline

\cite{xiao2025llm} &
MSDTMD-SG-Based Swarm Planning &
Planning / Path Optimization &
UAV Swarm Path Planning for Industry~5.0 Rescue Scenarios &
Learns human-like multi-step reasoning for movement decisions via fine-tuned datasets and symbolic modeling &
Highest task completion rate compared to DRL and ACO baselines with competitive latency and improved interpretability \\
\hline

\cite{wang2025multi} &
LLM-assisted Multi-UAV Placement &
Network-Level Planning &
Multi-UAV IAB Network Deployment &
Iteratively generates and refines UAV placement using structured prompting and historical feedback &
Near-optimal connectivity and coverage with significantly fewer iterations than conventional optimization methods \\
\hline

\cite{yuan2025next} &
NELV &
End-to-End Planning Pipeline &
Natural Language--Driven UAV Mission Execution &
Maps language instructions to executable flight trajectories via parsing, route planning, control, and monitoring &
Reduces pilot workload and enables multi-UAV missions while satisfying regulatory and environmental constraints \\
\hline

\cite{lykov2024flockgpt} &
FlockGPT &
Execution / Formation Control &
Natural Language--Driven UAV Swarm Formation &
Translates high-level geometric commands into executable formation control code using SDFs &
Enables intuitive swarm control for non-experts; effective formation execution without manual programming \\
\hline

\cite{wang2025rally} &
RALLY &
Execution / Hybrid Control &
Collaborative UAV Swarm Navigation &
Combines LLM reasoning with MARL via role assignment and goal refinement &
Improves task success rate, convergence speed, and robustness over MARL-only baselines \\
\hline

\cite{han2025swarmchain} &
SwarmChain (CoLLM) &
Execution / System Infrastructure &
On-Swarm LLM Inference for UAV Swarms &
Distributes LLM inference across UAVs using tensor parallelism and adaptive load scheduling &
Achieves 1.9--2.3$\times$ inference speedup and reduced latency without cloud dependence \\
\hline

\cite{chen2023typefly} &
TypeFly &
Execution / Low-Latency Control &
Real-Time LLM-Assisted UAV Control &
Generates control logic in MiniSpec with stream interpreting for immediate execution &
Up to 62\% reduction in response time and improved real-time responsiveness \\
\hline

\end{tabular}%
}
\end{table*}

\subsection{LLM-Assisted UAV Swarm Control and Execution}

UAV swarms have strong potential for collaborative, distributed missions, but their practical deployment poses substantial challenges at the execution and control layers. In addition to determining high-level plans, swarm systems must translate abstract objectives into real-time control actions, maintain formation and safety, coordinate multiple agents, and respond to environmental uncertainty under strict latency and resource constraints. These challenges are further amplified in scenarios where reliable communication with ground infrastructure is limited or unavailable.
\par
In this context, LLMs have emerged as a powerful interface connecting human intent, high-level reasoning, and low-level swarm execution. By enabling natural-language interaction, structured reasoning, and real-time code or policy generation, LLMs offer more intuitive and flexible swarm control than traditional interfaces. Unlike planning-oriented approaches, execution-level frameworks emphasize how swarm behaviors are generated, deployed, and coordinated in real-time, accounting for factors such as inference latency, distributed computation, and onboard resource limitations.
This subsection focuses on LLM-assisted swarm control and execution mechanisms, including natural-language-based formation control, hybrid LLM–RL architectures, distributed LLM inference across multiple UAVs, and low-latency code generation for control. These approaches position LLMs not only as planners but also as integral components of the control and system stack, enabling scalable, responsive, and human-centric swarm operation.
\par
Recent research examines how LLMs can improve UAV swarm control by enabling more intuitive and efficient human–swarm interaction. FlockGPT \cite{lykov2024flockgpt} is an early example of natural language–driven swarm formation control. Users provide high-level geometric commands, such as “form a sphere,” which GPT-4 translates into Python code using Signed Distance Functions (SDFs). These SDFs specify target UAV positions, while a flocking algorithm maintains cohesion and prevents collisions. User studies indicate that even non-experts can successfully command and interpret complex three-dimensional formations, demonstrating the accessibility of LLM-assisted swarm interfaces.
\par
Building on this language-driven paradigm, Role-Adaptive LLM-Driven Yoked Navigation for Agentic UAV Swarms (RALLY) \cite{wang2025rally} introduces a more structured framework that combines LLM reasoning with Multi-Agent RL (MARL). Unlike FlockGPT’s shape-centric control, RALLY focuses on collaborative navigation in dynamic, uncertain environments. It uses a two-stage LLM reasoning pipeline to interpret and refine natural-language goals. Its Role-value Mixing Network (RMIX) dynamically assigns roles – Commander, Coordinator, and Executor – balancing offline LLM knowledge with online MARL policies. This hybrid design improves coordination efficiency, task success rates, convergence speed, and robustness.
\par
While FlockGPT and RALLY focus on high-level planning, SwarmChain \cite{han2025swarmchain} addresses a salient systems challenge: LLM inference on resource-constrained UAVs. SwarmChain introduces the CoLLM framework, which distributes LLM computation across multiple UAVs using tensor parallelism and adaptive load scheduling, enabling real-time, on-swarm inference. A control layer employs dynamic prompt templates, vector databases, and context tracking to convert natural-language commands into executable swarm behaviors. Experiments report a 1.9–2.3× reduction in inference latency compared to existing distributed LLM approaches.
\par
In addition, Flying Drones with LLM (TypeFly) \cite{chen2023typefly} addresses latency by reducing the time required for LLMs to generate control code. Instead of using verbose languages such as Python, TypeFly introduces MiniSpec, a compact, token-efficient planning language designed for LLM-assisted control. Features such as skill probes, replanning, and stream interpretation – in which actions are executed as code is generated – reduce response time by up to 62\%. This significantly improves real-time responsiveness in time-critical UAV operations. Table \ref{tab:llm_uav_planning_control} provides a summary of these works.

Overall, the above studies demonstrate the rapid evolution of LLM-assisted UAV swarm intelligence. Each addresses a distinct challenge, including natural language interaction, collaborative reasoning, distributed inference, and low-latency control. Together, they indicate a future in which UAV swarms can be commanded and coordinated via scalable, intelligent, and seamless LLM-assisted interfaces.

\begin{table*}[t]
\centering
\caption{LLM-enhanced frameworks for UAV safety, network optimization, and autonomous intelligence, comparing methodologies, core LLM roles, and performance improvements in accident investigation, integrated communications, multi-objective optimization, open-world autonomy, and traffic management.}
\label{tab:llm_uav_safety_networks}
\resizebox{\textwidth}{!}{%
\begin{tabular}{|m{1cm}|m{2.4cm}|m{2.5cm}|m{4.0cm}|m{4.0cm}|m{4.0cm}|}
\hline
\textbf{Reference} &
\textbf{Proposed Framework} &
\textbf{Primary Application Domain} &
\textbf{Core Function of the LLM} &
\textbf{Key Advantage over Traditional Methods} &
\textbf{Key Performance Results} \\
\hline
\cite{yan2025uav} &
LLM--HFACS UAV Accident Investigation System &
UAV Safety Analysis and Accident Investigation &
LLM reasons over unstructured incident narratives and maps findings to the structured HFACS~8.0 taxonomy to identify causal factors across multiple levels &
Automates and standardizes human-factor analysis; detects subtle latent errors with reduced time, cost, and human effort compared to manual expert coding &
Macro-F1 scores of 0.58--0.76 across 18 categories; best F1 of 0.76 (precision 0.71, recall 0.82); some category accuracies $>$93\% \\
\hline
\cite{liu2025enhancing} &
LLM-assisted Satellite--UAV--IoT Network Optimization &
6G Integrated Satellite--UAV--IoT Communications &
Fine-tuned Llama~3~70B dynamically optimizes device association, resource allocation, and UAV positioning using real-time network telemetry &
Reduces action-space complexity and decision latency; enables adaptive, predictive network management beyond static optimization and RL methods &
27\% average spectral efficiency gain; 35\% reduction in decision latency compared to traditional approaches \\
\hline
\cite{li2025large} &
LEDMA (LLM-Assisted Decomposition-Based MOEA) &
UAV Networks with ISAC &
LLM acts as a black-box search operator to decompose and solve multi-objective optimization problems for UAV deployment and power control &
Improves convergence and solution diversity for non-convex, high-dimensional optimization beyond classical evolutionary algorithms &
Better Pareto front, faster convergence, and higher hypervolume than traditional multi-objective evolutionary algorithms \\
\hline
\cite{zhao2025general} &
Embodied Aerial Intelligent Agent &
Open-World Autonomous UAV Operations &
Edge-deployed LLM performs high-level task planning and scene understanding, interfaced with low-level perception, mapping, and motion control &
Enables robust autonomy and generalization in communication-constrained environments through hardware--software co-design &
14B-parameter LLM achieves 5--6 tokens/s at 220W; validated across diverse real-world missions (inspection, exploration, monitoring) \\
\hline
\cite{atrouz2024towards} &
LLM-Assisted Unmanned Traffic Management (UTM) System &
UAV Traffic Management and Airspace Safety &
LLM analyzes real-time airspace data, checks compliance, and generates actionable alerts and recommendations for human operators &
Enhances situational awareness and decision support beyond rule-based monitoring systems &
Up to 91.7\% decision accuracy with an average response time of 5.7 seconds across evaluated scenarios \\
\hline
\end{tabular}%
}
\end{table*}

\subsection{Leveraging LLMs for UAV Safety, Autonomy, and Network Optimization}

LLMs are increasingly contributing to various aspects of UAV systems, including safety analysis, autonomous decision-making, communications, traffic management, and education. In safety-critical applications, LLMs can identify contributory factors across multiple causal layers of UAV accidents. In communications, they help optimize the performance of Space–Air–Ground Integrated Networks. LLMs can also be integrated into Unmanned Traffic Management (UTM) systems, where they assist human operators by analyzing real-time data, interpreting complex situations, and generating actionable recommendations. This support improves the reliability, consistency, and timeliness of operational decisions. Additionally, educational platforms that leverage LLMs enable hands-on training in navigation, prompt engineering, and security scenarios, effectively bridging the gap between theoretical knowledge and practical UAV operations.
\par
In the initial research, Yan \textit{et al.}\cite{yan2025uav} propose an LLM-assisted UAV accident investigation framework that integrates the Human Factors Analysis and Classification System (HFACS) 8.0 model to automate the analysis of unstructured narrative incident reports. By combining HFACS’s structured taxonomy with LLM reasoning, the system systematically identifies contributory factors across all causal levels, including unsafe acts, latent preconditions, supervisory failures, and organizational influences. The framework was evaluated on 200 UAV incident reports, which contained over 3,600 coded instances, and achieved macro-F1 scores ranging from 0.58 to 0.76 across 18 HFACS categories. The best-performing configuration achieved a macro-F1 score of 0.76, with precision of 0.71 and recall of 0.82, while some category-level accuracies exceeded 93\%, matching or surpassing human expert performance on specific tasks. By improving analytical consistency and uncovering subtle latent errors, this approach reduces time and human effort while enabling more comprehensive, data-driven safety improvements.
\par
In the context of communications and network optimization, Liu \textit{et al.}\cite{liu2025enhancing} introduce an LLM-assisted framework for optimizing integrated satellite–UAV–IoT networks in 6G environments. The framework uses a fine-tuned Llama 3 70B model to dynamically manage device association, resource allocation, and UAV positioning based on real-time inputs, including satellite telemetry, IoT device status, and network traffic. To ensure computational efficiency, the authors implement action-space reduction and predictive UAV deployment strategies that adapt to environmental conditions and network load. Experimental results show an average 27\% improvement in spectral efficiency and a 35\% reduction in decision latency compared to traditional approaches, highlighting the framework’s robustness in dense, dynamic urban scenarios.
\par
Similarly, Li \textit{et al.}\cite{li2025large} address multi-objective optimization in UAV networks equipped with Integrated Sensing and Communication (ISAC) capabilities. They propose an LLM-assisted decomposition-based multi-objective evolutionary algorithm (LEDMA) that balances communication and sensing performance. By decomposing complex, non-convex optimization problems into simpler subproblems and using LLMs as black-box search operators, the framework iteratively refines solutions to UAV deployment and power control problems. Numerical evaluations show that LEDMA achieves a superior Pareto front, faster convergence, and higher hypervolume than conventional algorithms, demonstrating the effectiveness of LLMs in solving complex wireless optimization problems.
\par
Beyond optimization and networking, Zhao \textit{et al.}\cite{zhao2025general} introduce an aerial intelligent agent that integrates LLM-assisted reasoning with UAV autonomy for open-world task execution. Through a hardware–software co-design, the system deploys a 14B-parameter LLM on an edge-optimized platform, achieving inference speeds of 5–6 tokens per second at a peak power consumption of 220 W. A bidirectional cognitive architecture connects high-level LLM planning with low-level modules for state estimation, mapping, obstacle avoidance, and motion planning. The system is validated across diverse mission-critical applications, including sugarcane monitoring, power grid inspection, mine tunnel exploration, and biological observation. Results demonstrate robust task planning, scene understanding, and generalization, establishing a practical paradigm for embodied aerial intelligence.
\par
LLMs have also been explored as decision-support tools in UTM systems. For example, Atrouz \textit{et al.}\cite{atrouz2024towards} investigate the integration of LLMs into UTM operations, where they assist human operators by monitoring airspace activity, verifying UAV compliance with remote identification and mission plans, and flagging potential violations. Across the evaluated use cases, the system achieved up to 91.7\% accuracy and an average response time of 5.7 seconds, demonstrating its potential to enhance safety and efficiency in increasingly congested airspaces.
\par
In addition, the educational domain has also benefited from LLM integration. That is, FLY-LLM Sim \cite{11105292} is a modular laboratory platform designed for hands-on training with LAUS. The platform includes modules for advanced aerial navigation, where an LLM dynamically selects path-planning algorithms based on natural-language commands, as well as modules for analyzing vision-based cyberattacks on UAV perception systems. By combining the AirSim simulation environment with ChatGPT-4.0 for natural-language command processing and code generation, the platform provides a cost-effective, practical learning experience. User evaluations confirm its effectiveness and highlight its value for mastering LLM–UAV integration. Table \ref{tab:llm_uav_safety_networks} provides a summary of these works.

\begin{figure*}[h]
  \centering
  \includegraphics[width=0.85\textwidth]{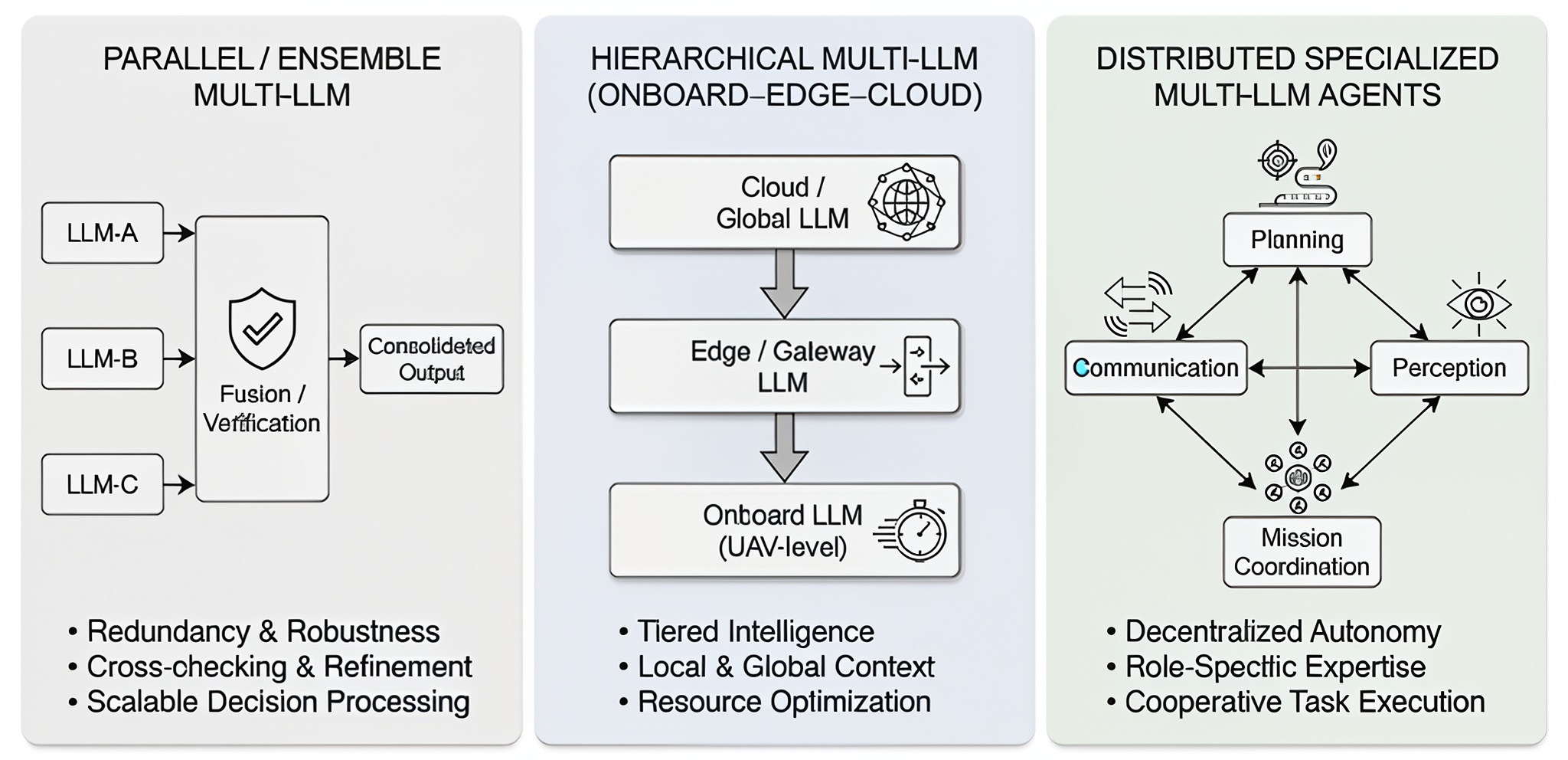} 
  \caption{Architectures for multi-LLM deployment in UAV swarms: parallel/ensemble for robust decision-making, hierarchical for tiered resource-aware intelligence, and distributed specialized agents for decentralized mission execution.}
  \label{fig:multillm}
\end{figure*}

\subsection{Multi-LLM Architectures for UAV Communications}

Single LLM agents are often unreliable for safety-critical systems and can generate biased or unsafe outputs. In UAV swarm operations, where real-time decisions have significant safety implications, relying on a single model increases the risk of errors or unsafe behavior. To address these risks, multiple LLMs can be used to improve reliability, reduce bias, and ensure decisions comply with operational constraints and safety regulations. Multi-LLM systems leverage the complementary strengths of individual models, allowing for specialization, scalability, and robustness. By employing task decomposition and cross-verification, these systems enhance overall decision quality and resilience, thereby making them more suitable for safety-critical UAV swarm applications.
\par
Multi-LLM systems are architectures in which multiple LLMs collaborate, either in parallel or sequentially, to solve complex tasks. Unlike single-model approaches, multi-LLM systems leverage the complementary strengths of individual models, enabling specialization, scalability, robustness, and effective task decomposition. Performance is further improved through cross-verification, which helps detect errors, reduce bias, and mitigate hallucinations. Collaboration among LLMs can follow cooperative, competitive, or ensemble paradigms, while system architectures may be centralized or decentralized, depending on fault-tolerance and task-allocation requirements. Additionally, multi-LLM frameworks can enhance privacy by routing sensitive queries to trusted models while validating outputs across agents.
\par
Deployment strategies differ significantly between cloud and edge environments. Edge deployments offer low latency and improved privacy by processing data locally, but are constrained by limited computational and energy resources. Therefore, techniques such as model compression, partitioning, and distributed inference are often necessary. In contrast, cloud deployments provide abundant computational capacity and support large-scale reasoning, but may experience higher latency and be less suitable for real-time or privacy-sensitive UAV operations.
\par
In aerial and UAV networks, multi-LLM systems enable distributed coordination among heterogeneous agents, including delivery UAVs, urban air mobility vehicles, and UAV-assisted wireless nodes. Individual LLMs can specialize in tasks such as route planning, communication optimization, or mission-level decision-making, while sharing critical information on navigation, sensing, and system status. This division of labor improves safety, efficiency, and resilience by supporting real-time adaptation, collision avoidance, fault tolerance, and collective optimization – capabilities that are difficult to achieve with a single LLM or a fully centralized controller \cite{luo2025toward}.
\par
Building on this paradigm, Cui \textit{et al.}\cite{cui2025research} propose a UAV swarm inspection system in which multiple lightweight LLM sub-models operate in parallel across a UAV fleet. Local LLMs handle onboard path planning and real-time decision-making, while a centralized orchestrator performs global reasoning, task allocation, and swarm-level optimization. By fusing multimodal data, including maps, telemetry, and visual and thermal imagery, the system dynamically adapts inspection paths and detects defects with high accuracy. Experimental results show improvements in coverage, energy efficiency, and robustness compared to traditional planning methods.
\par
Complementing this approach, Liu \textit{et al.}\cite{liu2025llm} introduce a multi-agent LLM framework tailored to UAV-based inspection in the Architecture, Engineering, and Construction (AEC) domain. Their system comprises specialized agents, including Router, PathPlanner, Controller, Perceptioner, and Retriever, that collectively support spatial reasoning, UAV control, and task delegation. A novel image-driven pipeline converts inspection data into semantic point clouds and three-dimensional scene graphs, enabling UAVs to interpret environments in alignment with human intent. Experiments demonstrate improved flexibility and automation in inspection workflows.
\par
Similarly, emphasizing distributed intelligence, Dharmalingam \textit{et al.} \cite{dharmalingam2025aero} present Aero-LLM, a hierarchical LLM architecture spanning the onboard, edge, and cloud layers. Each layer hosts specialized, fine-tuned models responsible for tasks such as anomaly detection\cite{emami2014efficient}, forecasting, and cybersecurity. This hierarchical decomposition enables low-latency responsiveness while supporting high-level analytical reasoning, all within lightweight computational footprints. Evaluations highlight Aero-LLM’s effectiveness in threat detection and secure decision-making on resource-constrained UAV platforms.
\par
Extending multi-LLM coordination to interconnected aerial networks, Yan \textit{et al. }\cite{yan2025hierarchical} propose a hierarchical control framework for hybrid terrestrial–non-terrestrial systems incorporating High-Altitude Platform Stations (HAPS). Specifically, a meta-controller LLM deployed on the HAPS manages global resource allocation and load balancing, while onboard LLMs handle real-time motion planning and communication decisions. Experimental results show that this collaborative architecture achieves higher system rewards, lower operational costs, and lower collision rates than DRL-assisted and hybrid baselines. Fig. \ref{fig:multillm} depicts Multi-LLM architectures for UAV communications, and Table \ref{tab:multi_llm_uav_networks} provides a summary of the works.

\par
Overall, distributed, cooperative, and domain-specialized LLM architectures represent a promising direction for the future of UAV systems. Multi-agent LLM frameworks enhance reliability, safety, and operational security by enabling robust decision fusion, coordinated resource management, synchronized mission planning, and decentralized control. These capabilities support scalable swarm coordination, real-time decision-making, and adaptive responses in dynamic environments, making multi-LLM systems well-suited for critical applications such as surveillance, disaster response, and infrastructure inspection, where fault tolerance and efficiency are essential.

\begin{table*}[t]
\centering
\caption{Hierarchical and multi-LLM architectures for coordinated UAV and aerial network operations, comparing distributed, swarm-based, and layered LLM frameworks in inspection, security, and hybrid terrestrial--non-terrestrial network applications.}
\label{tab:multi_llm_uav_networks}
\resizebox{\textwidth}{!}{%
\begin{tabular}{|m{1cm}|m{2.5cm}|m{2.5cm}|m{4.0cm}|m{4.0cm}|m{4.0cm}|}
\hline
\textbf{Reference} &
\textbf{Proposed Framework} &
\textbf{Primary Application Domain} &
\textbf{Core Function of the LLM} &
\textbf{Key Advantage over Traditional Methods} &
\textbf{Key Performance Results} \\
\hline
\cite{luo2025toward} &
Multi-LLM Distributed Coordination Paradigm &
Heterogeneous Aerial and UAV Networks &
Multiple specialized LLMs collaboratively handle routing, communication optimization, collision avoidance, and mission-level coordination while sharing critical state information &
Enables division of labor, real-time adaptation, fault tolerance, and collective optimization beyond single-LLM or centralized controllers &
Improved safety, efficiency, resilience, and scalability in heterogeneous aerial networks (conceptual and system-level validation) \\
\hline
\cite{cui2025research} &
Multi-LLM UAV Swarm Inspection System &
UAV Swarm Infrastructure Inspection &
Lightweight onboard LLMs perform local path planning and real-time decisions, while a centralized LLM orchestrator handles global reasoning and task allocation &
Balances local responsiveness with global optimization; improves robustness, coverage, and energy efficiency compared to traditional planners &
Higher inspection coverage, improved energy efficiency, and more robust swarm coordination in dynamic inspection scenarios \\
\hline
\cite{liu2025llm} &
LLM-assisted Multi-Agent Inspection Framework &
AEC Industry UAV Visual Inspection &
Specialized LLM agents (Router, PathPlanner, Controller, Perceptioner, Retriever) perform 3D reasoning, task delegation, perception, and UAV control &
Enables spatial-semantic understanding via semantic point clouds and 3D scene graphs; improves automation over rule-based inspection systems &
Demonstrates flexible task coordination and improved automation of complex inspection workflows \\
\hline
\cite{dharmalingam2025aero} &
Aero-LLM &
Secure and Resilient UAV Operations &
Hierarchical LLMs across onboard, edge, and cloud layers perform anomaly detection, forecasting, and cybersecurity analysis &
Achieves low-latency response and high-level reasoning while maintaining lightweight onboard computation &
Strong performance in threat detection and secure decision-making on resource-constrained UAV platforms \\
\hline
\cite{yan2025hierarchical} &
Hierarchical LLM-Controlled Hybrid Aerial Network &
UAV--HAPS Terrestrial--Non-Terrestrial Networks &
Meta-controller LLM on HAPS manages global resource allocation and load balancing, while onboard LLMs handle real-time motion and communication control &
Joint optimization of motion and communication outperforms DRL and hybrid control schemes in complex aerial networks &
Higher system rewards, lower operational cost, and reduced collision rates compared to DRL and hybrid baselines \\
\hline
\end{tabular}%
}
\end{table*}

\subsection{Key Findings and Insights}

Based on the above subsections, here we discuss the key findings and insights.

\subsubsection{\textbf{LLM-assisted Spectrum Management}}

Recent frameworks for LLM-assisted green-spectrum management in 6G Non-Terrestrial Networks (NTNs) integrate LLMs into a three-layer hierarchical architecture comprising data collection, intelligent LLM processing, and feedback-enabled execution. This design enables dynamic and adaptive spectrum allocation across satellites, high-altitude platforms, UAVs, and terrestrial networks. A key innovation is the joint optimization of spectral efficiency and energy consumption, supported by mechanisms including real-time interference analysis, predictive energy modeling, and an explicit energy–spectrum trade-off function. Simulation results show that the LLM-assisted approach significantly outperforms conventional methods, supporting up to 175 active transmitters – a 167\% increase in capacity – while maintaining stable energy consumption and tighter interference control under varying threshold conditions. These findings advance intelligent and sustainable spectrum management for future 6G NTNs and underscore the transformative potential of LLMs in enabling green, scalable, high-performance global communications \cite{rong2025llm}.

\subsubsection{\textbf{LLM-assisted Near-Field Communications}}

The integration of LLMs into near-field communications for the Low-Altitude Economy (LAE) brings transformative capabilities that address the unique challenges of UAV-centric wireless networks. By analyzing high-dimensional channel-state information in real time, LLMs can intelligently classify users as near-field or far-field without requiring precise distance estimation, thereby reducing computational overhead. They also enhance multi-user precoding by learning to predict beamforming and power allocation parameters directly from channel data, enabling efficient real-time adaptation to dynamic UAV mobility. Leveraging strong feature-extraction and pattern-recognition capabilities, LLMs can process the massive amounts of antenna and user data generated by extremely large-scale Multiple-Input Multiple-Output (MIMO) systems, thereby enabling performance to scale effectively as network density increases. In addition, their adaptability and few-shot learning abilities enable generalization across diverse operating conditions, such as varying UAV densities or interference levels, without extensive retraining. Finally, LLMs enable unified multi-task learning by jointly handling user classification, precoding, and interference management within a single model, thereby significantly simplifying the overall system design \cite{xu2025empowering}.

\subsubsection{\textbf{LLM-assisted ISAC Optimization}}

The integration of LLMs into the optimization of Integrated Sensing and Communication (ISAC) systems represents a significant methodological advance, as demonstrated by two recent studies that employ LLMs in complementary roles in complex wireless network design. Li \textit{et al.}\cite{10839306} examine a multi-UAV ISAC network in which UAVs must jointly provide communication services and radar-based localization of ground users. In this study, an LLM (GPT-3.5 Turbo) is incorporated as an intelligent search operator within a Multi-Objective Evolutionary Algorithm (MOEA). By prompting the LLM with the problem context and current solution states, the model generates novel candidate solutions for UAV deployment and power allocation. This LLM-assisted search enables the evolutionary algorithm to more effectively explore the non-convex trade-off between network utility and sensing accuracy, resulting in improved Pareto fronts and faster convergence compared to conventional approaches.
\par
Complementarily, Li \textit{et al.}\cite{li2025joint} study a multi–base station MIMO ISAC system and address the joint optimization of discrete user association and continuous transmit beamforming. In this framework, the LLM (GPT-o1) serves as a combinatorial solver for the NP-hard user-association subproblem within an alternating-optimization loop. Through carefully engineered prompts incorporating ICL, CoT reasoning, and iterative self-reflection, the LLM infers high-quality association strategies. These results are then combined with rigorous convex optimization techniques—such as fractional programming and the Alternating Direction Method of Multipliers (ADMM), for beamforming design, yielding a hybrid solution that achieves near-optimal performance, closely matches brute-force upper bounds, and outperforms conventional game-theoretic methods.

\subsubsection{\textbf{High-Speed Connectivity for LLM-assisted UAVs}}

The effectiveness of LLM-assisted reasoning in UAV systems depends critically on the reliability, latency, and timeliness of the underlying communication links. While LLMs extend UAV capabilities beyond automation by supporting summarization, semantic understanding, and adaptive mission support, these functions rely on continuous, high-quality data exchange. UAV-assisted URLLC therefore serves as a foundational layer that enables closed-loop perception, real-time human–machine interaction, and dependable reasoning in mission-critical scenarios such as disaster response, search and rescue, and infrastructure inspection. Without ultra-reliable and low-latency communication, the system-level advantages of LLM-assisted intelligence cannot be sustained in real-world deployments.

5G networks, leveraging key technologies such as Millimeter-Wave (mmWave) communication, massive MIMO, and advanced beamforming, enable URLLC and are well-suited for mission-critical applications such as UAV-assisted SAR. URLLC can achieve end-to-end latencies of 1 ms or less, supporting real-time decision-making in highly time-sensitive scenarios. The resulting high-throughput, low-latency communication links between UAVs and edge infrastructure allow rapid transmission of multimodal sensor data for immediate processing. This capability enables on-the-fly LLM inference, allowing UAVs to make responsive, context-aware decisions in dynamic, uncertain environments. By offloading computationally intensive tasks to edge servers, 5G connectivity effectively overcomes many of the resource constraints inherent to onboard UAV platforms.


\subsubsection{\textbf{Assessing LLM Reasoning and Safety}}

Field deployments of integrated LLM-enabled UAV systems confirm significant improvements in operational adaptability and effectiveness across diverse applications, including emergency response, environmental monitoring, and infrastructure inspection. At the same time, these evaluations reveal persistent constraints in bandwidth availability, energy consumption, multimodal robustness, and reasoning reliability. These observations highlight the necessity of holistic co-design across communication, computation, and control, as well as the importance of standardized evaluation frameworks, such as UAVBench and UAVThreatBench, to objectively assess reasoning performance and cybersecurity readiness for safe and scalable deployment of intelligent UAV communication systems.
The benchmark introduced in \cite{ferrag2025uavbench} addresses this gap by providing 50,000 validated UAV scenarios and an additional 50,000 multiple-choice questions ($UAVBench_{MCQ}$) spanning 10 distinct reasoning styles. This comprehensive design enables systematic evaluation of a wide range of cognitive capabilities.

\section{MLLMs: Integrating Vision, Language, and Reasoning in UAV Operations and Communications} 
\label{sec5}
This section examines the emergence of MLLMs as a unifying intelligence paradigm that integrates visual, linguistic, and sensor information for UAV systems. By extending large language models with multimodal perception and reasoning capabilities, MLLMs enable UAVs to interpret complex environments, understand high-level human intent, and support context-aware autonomous decision-making.

\subsection{Building and Fine-Tuning Multimodal LLMs}

Training MLLMs for UAV applications begins with constructing multimodal datasets that reflect the unique sensing conditions of aerial platforms. Typical data sources include aerial images and videos, onboard sensor measurements, and associated textual descriptions or operational logs. Compared with ground-level imagery, UAV data exhibit distinctive viewpoints, scale variations, and occlusions, making UAV-specific datasets essential for effective multimodal alignment. Consequently, data augmentation and domain-aware preprocessing are commonly employed to improve robustness under diverse flight conditions.
Most MLLMs are initialized through large-scale self-supervised multimodal pretraining to acquire general cross-modal representations. While techniques such as contrastive learning are widely adopted, their primary role in UAV applications is to provide transferable multimodal priors rather than task-specific aerial knowledge. As a result, pretrained models often require substantial adaptation to handle the geometric and semantic characteristics of aerial scenes.

A key challenge lies in aligning heterogeneous modalities for grounded reasoning. Contemporary MLLMs typically rely on modality-specific encoders coupled with a pretrained LLM, thereby enabling language modules to selectively attend to visual features. This flexible architecture serves as a common backbone for UAV perception and reasoning tasks, but its effectiveness in aerial settings depends strongly on downstream adaptation.

Following pretraining, models are fine-tuned on UAV-specific tasks such as aerial image captioning, visual question answering, and object recognition. Dedicated datasets, including AeroCaps \cite{basak2025aerial} and HRVQA-VL \cite{guan2025uav}, support fine-grained scene understanding and multi-hop reasoning over UAV imagery. Empirical evidence shows that task-aware fine-tuning is critical: compact MLLMs (e.g., 2B parameters) adapted with UAV data can outperform significantly larger generic models while requiring only a few gigabytes of FP16 memory.

In practice, training large multimodal models from scratch is rarely feasible for UAV systems. Instead, transfer learning and parameter-efficient adaptation techniques are widely adopted. By freezing pretrained backbones and updating lightweight modules such as low-rank adapters, MLLMs can be efficiently aligned with aerial semantics at reduced computational and memory cost, making them suitable for edge-deployed and resource-constrained UAV platforms \cite{guan2025uav}.

Overall, effective MLLM deployment in UAV applications relies less on generic multimodal pretraining strategies and more on UAV-specific data, efficient adaptation, and careful alignment with operational constraints.

\begin{figure*}[t!]
  \centering
  \includegraphics[width=\textwidth]{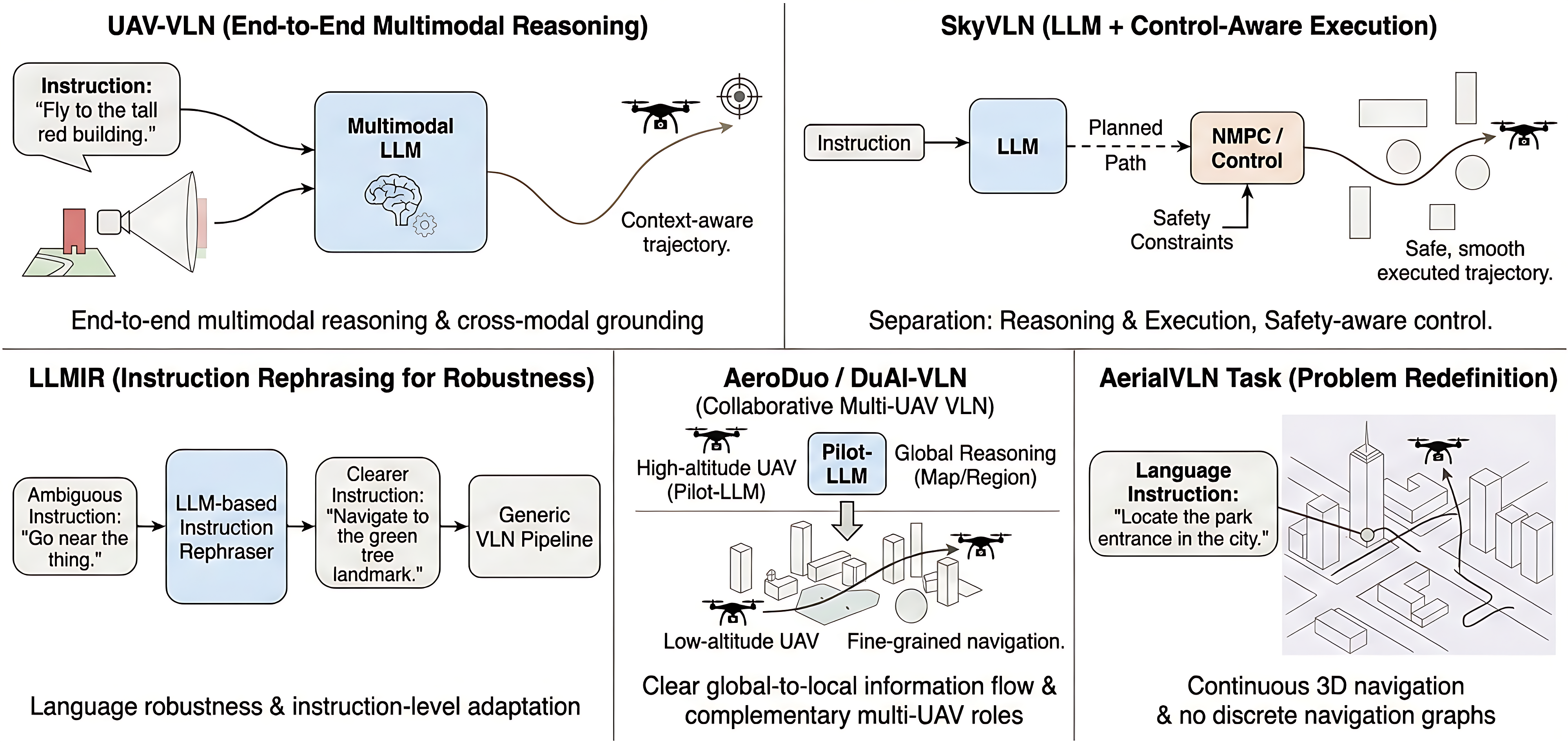} 
  \caption{End-to-end multimodal reasoning and control frameworks for VLN in UAVs, featuring instruction grounding, hierarchical collaboration, and safety-aware execution for continuous 3D navigation.}
  \label{fig:mllm}
\end{figure*}

\subsection{MLLM-Assisted Human–Swarm Interaction and Autonomy}

Integrating MLLMs into UAV systems enables UAVs to more effectively synthesize information from multiple sensors. In the following, we examine how MLLMs can enhance the performance and scalability of UAV systems, focusing on their potential to improve multi-agent perception, navigation, and coordination within UAV swarms, as well as their role in ensuring secure and resilient swarm operations.
\par
Recent advances in MLLMs have significantly expanded the capabilities of Human–Swarm Interaction (HSI) and autonomous UAV systems. Eunmi \textit{et al.}\cite{eumi2025swarmchat} highlight this shift by introducing SwarmChat, a context-aware interaction framework in which LLM-assisted modules, context generation, intent recognition, task planning, and modality selection collaborate to interpret user commands and adapt them to real-time swarm states. Their results demonstrate that natural-language, voice, and teleoperation interfaces can substantially reduce cognitive load. However, the system’s rule-based intent recognition and limited experimental scope indicate room for more scalable, learning-based solutions. Building on the idea that LLMs can mediate complex swarm behaviors, Liu \textit{et al.}\cite{liu2024multi} present a multimodal GPT-4–based approach for UAV formation control using both visual and textual inputs. Their results show that MLLMs can extract commands from human instructions, interpret environmental cues from onboard imagery, and plan formations with over 80\% success. This work validates the feasibility of end-to-end multimodal reasoning for swarm perception, coordination, and adaptive task execution, complementing the interaction-focused perspective introduced by SwarmChat. Extending multimodal reasoning into broader autonomous decision-making, Ping \textit{et al.}\cite{ping2025multimodal} explore how MLLMs can enhance multi-agent perception, navigation, and coordination within UAV swarms. Through a forest-firefighting case study, they demonstrate that an MLLM (Doubao 1.5 Vision Pro) can perform natural-language task planning, dynamic fire assessment, and coordinated execution across multiple UAVs. Their discussion of challenges, including hallucinations, inference latency, and communication bottlenecks, further contextualizes the limitations identified in earlier work and underscores the need for more efficient, robust multimodal pipelines.

\begin{table*}[t!]
\centering
\caption{Multimodal LLM applications in UAV swarm operations, comparing frameworks for human-swarm interaction, autonomous formation, threat detection, natural language control, and emergency monitoring across diverse sensory and linguistic inputs.}
\label{tab:multimodal_llm_uav}
\begin{tabular}{|m{1cm}|m{2.5cm}|m{2.5cm}|m{4cm}|m{6cm}|}
\hline
\textbf{Reference} & \textbf{Purpose} & \textbf{Input Modalities} & \textbf{Models} & \textbf{Key Outcomes / Achievements} \\
\hline
\cite{eumi2025swarmchat} & Human-swarm interaction, context-aware command & Text, voice, teleoperation & LLM-assisted modules: Context Generator, Intent Recognition, Task Planner, Modality Selector & Reduces cognitive load, interprets commands accurately, provides flexible communication modes; requires further learning-based intent understanding and large-scale validation \\
\hline
\cite{liu2024multi} & Autonomous swarm formation, safe command execution & Images, text & GPT-4 multimodal LLM, pre-training on single-UAV & Achieves 82.7\% command extraction accuracy; 83.8\% formation planning success; handles ambiguous human instructions, integrates perception and swarm status \\
\hline
\cite{ping2025multimodal} & Dynamic mission planning, multi-agent coordination, situational awareness & Text, visual, sensor data & MLLM (Doubao 1.5 Vision Pro), chain-of-thought reasoning, multi-source data fusion & Autonomous swarm management, adaptive task execution, natural language interaction; addresses hallucination and inference-latency trade-offs \\
\hline
\cite{xiao2024multi} & Cyber-physical threat detection in UAV swarms & Sensor data, logs, UAV communications & Large multimodal models, hierarchical zero-trust architecture, edge-fog-cloud scheduling & Improves robustness and timeliness in detecting threats; balances inference accuracy, latency, and computational constraints \\
\hline
\cite{11213101} & Intuitive UAV control via natural language & Text, video, image & MLLMs (Gemini / LLaVA), logistic regression classifier, YOLOv10, DeepSORT, MAVSDK & High-accuracy command execution (97.58\% efficiency in video selection), effective autonomous operation, reduces need for specialized programming skills \\
\hline
\cite{11088533} & Real-time fire monitoring, automated report generation & UAV imagery, ground sensors, meteorological data & LLM (DeepSeek), Grounding DINO, YOLOv8, improved A* path planning & Generates combined text-and-image reports, identifies fireground elements, optimizes evacuation routes, enhances situational awareness and operational efficiency \\
\hline
\end{tabular}
\end{table*}

In addition to these operational advances, Xiao \textit{et al.} \cite{xiao2024multi} examine the role of MLLMs in secure and resilient swarm operations and propose a multimodal intrusion-detection architecture that integrates cyber and physical data streams. Their hierarchical zero-trust framework distributes models of varying sizes across edge, fog, and command layers, optimizing inference latency and computational cost while enhancing detection robustness. This security-focused perspective underscores the need for trustworthy MLLM integration as swarms become increasingly autonomous and data-intensive. From a control interface standpoint, Balaji \textit{et al. }\cite{11213101} introduce PromptPilot, which uses natural language instructions to control UAVs without requiring domain expertise. By combining a lightweight intent classifier with MLLMs such as Gemini or LLaVA, along with YOLOv10 and DeepSORT for vision-aware tasks, PromptPilot generates executable control code and demonstrates reliable autonomous performance. This work aligns with SwarmChat and Liu \textit{et al. }in showing that intuitive human–UAV interaction can be achieved through MLLMs, while emphasizing practical deployment and real-world testing. Finally, Yang \textit{et al.}\cite{11088533} focus on rapid multimodal situational reporting, integrating UAV imagery, ground sensor data, and meteorological information to generate comprehensive forest-fire intelligence. Their system fuses Grounding DINO, YOLOv8-based models, and the DeepSeek LLM to identify critical environmental features and compute safe evacuation routes via enhanced A. This application illustrates how multimodal perception and language-based reasoning can support high-stakes environmental decision-making, complementing the swarm-control and interaction frameworks previously discussed. Table \ref{tab:multimodal_llm_uav} provides a summary of these works.

\par
Overall, SwarmChat integrates modules for context generation, intent recognition, task planning, and modality selection, enabling UAV swarms to interpret user commands and adapt to real-time operational states. Experiments show that natural language, voice, and teleoperation interfaces can substantially reduce cognitive load. However, current rule-based intent recognition and limited testing highlight the need for more scalable, learning-based solutions. Building on this approach, PromptPilot enables UAV control via natural-language instructions without requiring domain expertise. By combining a lightweight intent classifier with MLLMs such as Gemini or LLaVA and vision-aware models such as YOLOv10 and DeepSORT, PromptPilot translates user instructions into executable control code and demonstrates reliable autonomous performance. Together, these systems illustrate the potential of MLLMs to mediate complex swarm behaviors, enhance human-machine interaction, and improve the efficiency, scalability, and resilience of UAV operations.

\subsection{MLLM-Assisted UAV Navigation and Collaboration}

UAV navigation is increasingly framed as a multimodal reasoning problem that requires the tight integration of language understanding, visual perception, and control. Recent advances in VLN reflect this shift, with emerging frameworks enabling UAVs to interpret free-form instructions, reason over complex visual scenes, and execute context-aware flight in previously unseen environments. Collectively, these developments have led to substantial improvements in instruction-following accuracy, generalization, and robustness, accelerating progress toward language-driven UAV autonomy.

Saxena et al. \cite{saxena2025uav} introduce UAV-VLN, an end-to-end framework that reformulates aerial navigation as a question-answering problem. By combining LLM-based commonsense reasoning with open-vocabulary visual perception, UAV-VLN grounds natural-language instructions in real-time visual observations to generate navigation trajectories (Fig. \ref{fig:mllm}). Experimental results demonstrate strong zero-shot generalization across diverse indoor and outdoor environments, highlighting the effectiveness of cross-modal grounding for aerial VLN.

Extending VLN toward safe and reliable execution, Li et al. \cite{li2025skyvln} propose SkyVLN, which integrates LLM-based instruction reasoning with a safety-aware Nonlinear Model Predictive Control (NMPC) module (Fig. \ref{fig:mllm}). By incorporating fine-grained spatial verbalization and path-history memory, SkyVLN improves robustness to ambiguous instructions while enforcing collision avoidance and dynamic constraints during flight. Simulation results show that coupling high-level language reasoning with model-based control enhances both task success and operational safety.

Rather than modifying navigation or control directly, Chen et al. \cite{chen2023vision} focus on improving language robustness in UAV navigation. Their LLM-assisted Instruction Rephraser (LLMIR) reformulates vague or out-of-distribution user commands into clearer, model-compatible instructions (Fig. \ref{fig:mllm}). This preprocessing strategy significantly improves navigation success under challenging linguistic conditions, offering a practical solution for user-adaptive and reliable human–UAV interaction.

Moving beyond single-agent settings, Wu et al. \cite{wu2025aeroduo} extend VLN to collaborative scenarios through DuAl-VLN, a dual-UAV navigation framework operating at complementary altitudes. Their AeroDuo system assigns global reasoning and target localization to a high-altitude UAV equipped with a multimodal LLM, while a low-altitude UAV performs fine-grained, collision-free navigation using structured path planning (Fig. \ref{fig:mllm}). Supported by the HaL-13k dataset, AeroDuo significantly outperforms single-UAV baselines, demonstrating the advantages of cooperative reasoning in complex aerial environments.

At a broader level, Liu et al. \cite{liu2023aerialvln} introduce AerialVLN, a benchmark task that redefines UAV navigation in open, continuous 3D urban spaces. Unlike ground-based VLN, AerialVLN requires UAVs to navigate freely using first-person visual input while handling altitude variation and ambiguous spatial references. This task formalizes key challenges in cross-modal grounding and 3D spatial reasoning that contemporary frameworks such as UAV-VLN, SkyVLN, and AeroDuo aim to address. A comparative summary is provided in Table \ref{tab:vln_frameworks_comparison}.

Overall, recent advances in VLN demonstrate that tightly integrating LLM-assisted reasoning with visual perception, control-aware execution, and multi-UAV collaboration enables UAVs to translate natural-language instructions into real-time, context-aware flight behavior. These developments establish a strong foundation for future research in scalable, language-driven aerial navigation.

\begin{table*}[t!]
\centering
\caption{Comparison of VLN frameworks for UAVs, highlighting innovations in end-to-end reasoning, safety-aware control, instruction robustness, multi-UAV collaboration, and benchmark development for continuous 3D aerial navigation.}
\label{tab:vln_frameworks_comparison}
\begin{tabular}{|m{1cm}|m{3.5cm}|m{3.5cm}|m{4cm}|m{4cm}|}
\hline
\textbf{Reference} & \textbf{Core Innovation / Focus} & \textbf{Key Technical Components} & \textbf{Key Outcomes / Performance} & \textbf{Limitations / Challenges Addressed} \\
\hline
\cite{saxena2025uav} & End-to-end VLN; reframes navigation as QA; cross-modal grounding for interpretability. & Fine-tuned LLM, open-vocabulary visual perception, novel cross-modal grounding mechanism. & Significant improvement in instruction-following accuracy \& path efficiency; robust zero-shot generalization in diverse indoor/outdoor settings. & Minimizes task-specific supervision; outperforms closed-vocabulary baselines. \\
\hline
\cite{li2025skyvln} & Integrates VLN with NMPC for safe, precise urban flight. & LLM, fine-grained spatial verbalizer, history path memory, NMPC module for control optimization. & Substantial improvement in navigation success rate \& operational efficiency; ensures safe, constraint-aware flight. & Reduces human operator cognitive load and advances explainable autonomous decision-making. \\
\hline
\cite{chen2023vision} & Addresses out-of-distribution (OOD) instructions via an LLM-assisted Instruction Rephraser (LLMIR). & Conditional Transformer (VLCT), SentenceBERT, LLM-assisted Instruction Rephraser (LLMIR). & 1.39\% absolute improvement in task success rate; +1.51\% improvement on OOD test set. & Enhances reliability with vague/user instructions; improves user-friendliness. \\
\hline
\cite{wu2025aeroduo} & Dual-UAV collaboration at different altitudes for superior navigation. & Multimodal LLM (Pilot-LLM), orthographic mapping, multi-stage pathfinding pipeline. & Significantly outperforms single-UAV methods; higher success rates \& better generalization on unseen environments/objects. & Overcomes limitations of single-UAV perspective via collaborative reasoning. \\
\hline
\cite{liu2023aerialvln} & Proposes a new VLN benchmark for UAVs in continuous 3D outdoor urban spaces. & Task definition for continuous, open 3D navigation with first-person visual perceptions. & Introduces challenges in 3D spatial reasoning, language understanding, and cross-modal alignment. & Addresses the gap in aerial-agent navigation for complex urban environments. \\
\hline
\end{tabular}
\end{table*}

\begin{figure}[t!]
  \centering
  \includegraphics[width=0.45\textwidth]{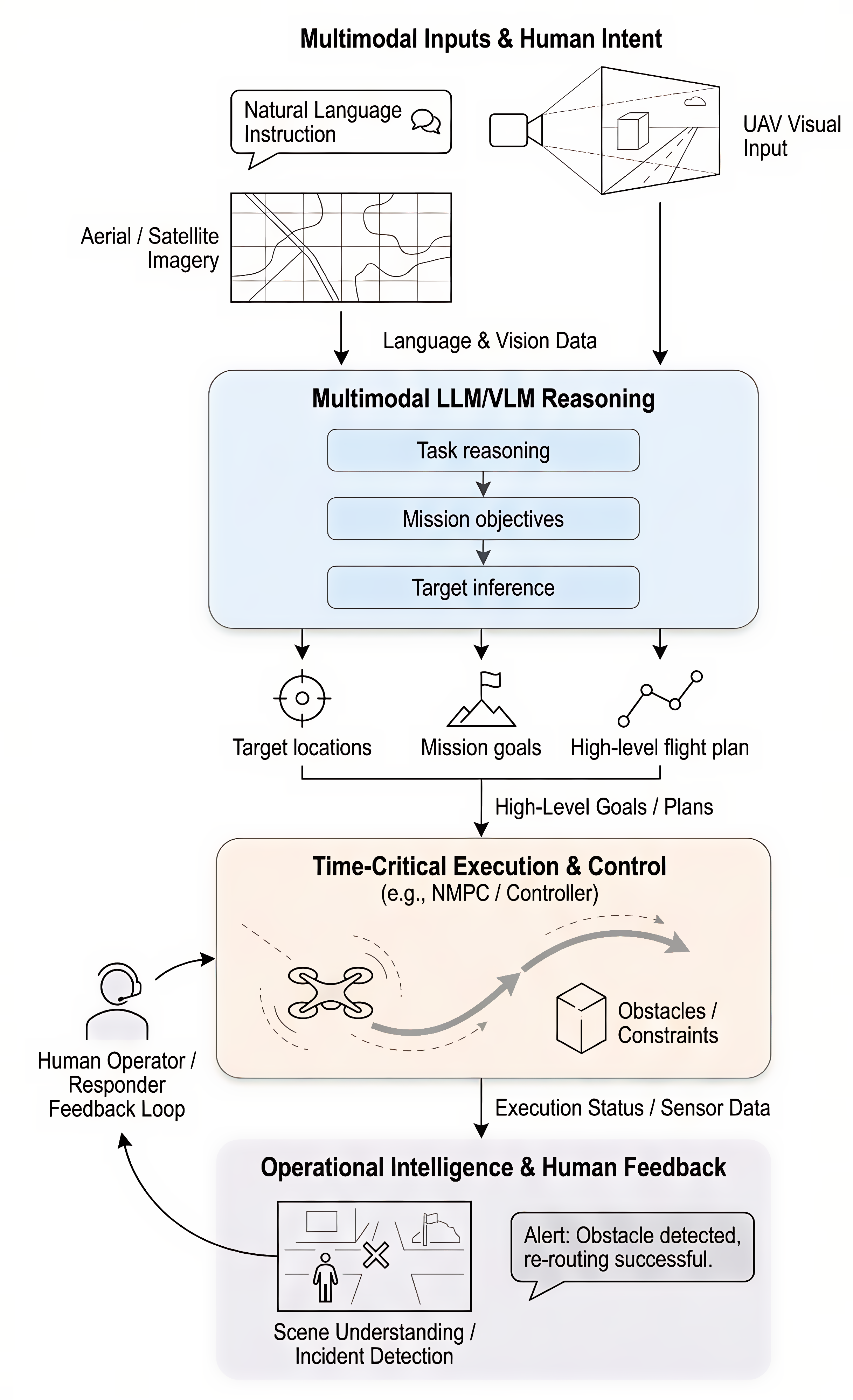} 
  \caption{Integrated high-level planning and operational intelligence pipeline for UAVs, using multimodal LLMs/VLMs to interpret instructions, infer goals, and generate flight plans, with real-time feedback loops for execution monitoring and situational awareness.}
  \label{fig:vllm}
\end{figure}

\subsection{LLM/VLM-Assisted High-Level Mission Planning, Task Reasoning, and Operational Intelligence}

The rapid advancement of multimodal AI is fundamentally transforming UAV capabilities, enabling aerial platforms to perceive complex environments, interpret natural language instructions, and autonomously execute mission-critical tasks. By tightly integrating LLMs with vision-based perception systems, modern UAVs can infer user intent, generate context-aware plans, and adapt dynamically to real-world conditions with minimal human intervention. These capabilities support a wide range of applications, including rapid SAR operations, strategic mission planning, real-time incident detection, and situational analysis. Multimodal pipelines that integrate high-precision object detection, scene-level reasoning, and natural language summarization enable UAVs to convert raw sensor data into actionable intelligence, thereby bridging the gap between human intent and autonomous aerial execution. The overall process is depicted in Fig. \ref{fig:vllm}.

\par
Recent research has increasingly focused on enabling UAVs to autonomously interpret visual scenes, understand natural language, and perform time-critical operations. Yaqoot et al. \cite{yaqoot2025uavvlrrvisionlanguageinformednmpc} exemplify this trend with UAV-VLRR, a rapid SAR framework that integrates a multimodal vision–language reasoning system using ChatGPT-4o and Molmo with a high-performance NMPC. By directly translating language commands and aerial imagery into actionable target coordinates and executing point-to-point NMPC control with integrated obstacle avoidance, UAV-VLRR eliminates the need for preplanned trajectories. Experimental results demonstrate significant speedups over both human pilots and conventional autopilots, highlighting the effectiveness of combining MLLMs with real-time control for agile, safety-critical missions.
\par
Complementing rapid execution at the control level, Sautenkov et al. \cite{sautenkov2025uav} extend multimodal reasoning to strategic mission planning with UAV-VLA. Instead of directly controlling low-level flight, their system enables users to specify large-scale aerial missions using natural language, which an LLM interprets as high-level objectives. A vision–language model then identifies relevant targets in satellite imagery, and a final LLM module generates a complete flight plan. Although the resulting trajectories are slightly longer than those designed by expert operators, they are produced more than six times faster, demonstrating the potential of MLLMs to significantly reduce operator workload and streamline global-scale mission design.
Beyond navigation and planning, multimodal AI has also demonstrated strong potential for real-time incident detection and situational awareness. Eesaar et al. \cite{11208573} present an autonomous UAV system that integrates a DJI Matrice 30T with an end-to-end multimodal pipeline for emergency response. In this system, YOLOv12n detects accidents and fires in real-time aerial video, LLaVA-OneVision-Qwen2 generates detailed scene-level descriptions, and GPT-4o mini produces concise natural-language summaries for first responders. Field validation shows that this pipeline can rapidly transform raw aerial imagery into actionable alerts, significantly improving response times and situational awareness in transportation and emergency management scenarios.
\par
Taken together, these works illustrate a growing ecosystem of multimodal UAV systems that translate vision and language understanding into meaningful action. From fast, reactive control in UAV-VLRR, to large-scale mission planning in UAV-VLA, to real-time scene interpretation for safety-critical response, recent advances highlight the expanding role of MLLMs in enabling intelligent, adaptive, and resilient aerial autonomy. As multimodal AI matures, it is set to play a central role in developing reliable, interpretable, and fully autonomous UAV operations across diverse real-world applications.

\subsection{LLM/VLM-Driven Autonomous Control and Real-Time Flight Decision-Making}

LLMs and VLMs in UAV systems pose a critical challenge: enabling autonomous agents to accurately infer human intent, interpret complex, dynamic environments, and make safe, context-aware decisions despite the uncertainty and potential hallucinations inherent in modern foundation models. 
\par
As multimodal large models continue to expand the capabilities of autonomous UAVs, recent research has increasingly focused on improving reliability, transparency, and resource efficiency in real-world deployment. Hu et al. \cite{hu2025llvm} address one of the most pressing challenges—hallucination in LLMs and VLMs—through LLVM-UAV, a framework that tightly couples a Domain-Guided Structured Prompt Execution Framework (DGSPEF) with lightweight, task-specific vision models. By translating natural-language instructions into structured, domain-constrained executable code and validating perception using independent visual modules, LLVM-UAV enables reliable zero-shot task execution without additional model training. Extensive hardware experiments and evaluations across eight LLMs demonstrate strong robustness, positioning the framework as a practical solution for safety-critical UAV applications such as mapping, disaster response, and precision agriculture.
\par
Building on the theme of reliability while emphasizing human collaboration, Krupavs et al. \cite{krupavs2025multimodal} introduce a multimodal architecture for human–machine collaborative UAV navigation. Their system uses GPT-4.1-nano for multimodal reasoning and generates navigation commands, natural-language explanations, and confidence scores, which are presented to operators via a touch-based approval interface. This design enables a closed-loop interaction paradigm in which operators can accept, reject, or modify model-generated actions. Experimental results show high levels of user trust and explainability, with approximately 90\% approval of AI-generated instructions. However, observed response latency and imperfect calibration between confidence scores and actual performance highlight ongoing challenges in deploying explainable VLMs for real-time aerial control.
\par
While LLVM-UAV and human–machine collaboration frameworks prioritize reliability and transparency, Wang et al. \cite{wang5397714lmucs} focus on enabling efficient onboard autonomy for resource-constrained UAV platforms. Their LMUCS framework integrates a compact, LoRA-tuned LLM for natural language parsing with advanced perception modules, including YOLOv11 for object detection and MiDaS for monocular depth estimation. Achieving over 94\% command-parsing accuracy and strong visual perception performance, LMUCS enables lightweight UAVs to execute complex missions, such as autonomous material search and delivery, without relying on cloud-based computation. This work demonstrates that high-level language reasoning and robust 3D perception can be jointly realized on embedded platforms.
\par
Extending these ideas to goal-directed navigation, Zhang et al. \cite{zhang2025grounded} propose VLFly, a unified vision–language navigation framework explicitly designed for UAVs that avoids the common reliance on discrete action spaces and external localization systems. Instead, VLFly directly outputs continuous velocity commands from monocular camera input, integrating an LLM-assisted instruction encoder, a VLM-based goal retriever, and a trajectory planner to generate real-time waypoints. Evaluated across diverse simulation environments without task-specific fine-tuning, VLFly consistently outperforms previous approaches. Real-world flight experiments in both indoor and outdoor settings further confirm its robust open-vocabulary grounding and its ability to interpret abstract or indirect language instructions, highlighting its suitability for zero-shot deployment in unconstrained UAV navigation scenarios. Table \ref{tab:uav_autonomy_challenges}
summarizes challenges and solution directions in LLM/VLM-Based UAV Autonomy.
\par
Integrating LLMs and VLMs into UAV platforms requires addressing challenges in intent understanding, environmental perception, and safe decision-making under model uncertainty. Recent approaches address these issues through structured prompt execution, lightweight task-specific perception, and explainable multimodal interaction. Efficient onboard autonomy frameworks demonstrate that compact language models and advanced visual perception can enable robust, real-time operation on resource-limited UAVs. These advances represent a significant step toward reliable, interpretable, and fully autonomous multimodal aerial systems.

\begin{table*}[t!]
\centering
\caption{Key challenges in LLM/VLM-based UAV autonomy—including model uncertainty, interpretability, resource constraints, environmental perception, and intent understanding—paired with solution approaches, required capabilities, and desired operational outcomes.}
\label{tab:uav_autonomy_challenges}
\begin{tabular}{|m{4cm}|m{4.8cm}|m{4cm}|m{4cm}|}
\hline
\makecell{\textbf{Core} \\ \textbf{Challenge}} &
\makecell{\textbf{Solution} \\ \textbf{Approach}} &
\makecell{\textbf{Key}  \textbf{Capability}} &
\makecell{\textbf{Desired} \textbf{Outcome}} \\
\hline
Model uncertainty and hallucinations of foundation models & Combining structured language reasoning with robust visual perception & Error-resilient, zero-shot task execution across diverse missions & Greater reliability and safety in real-world aerial missions \\
\hline
Lack of interpretability and trust in AI decisions & Pairing VLM-based reasoning with real-time depth perception and operator-guided approval loops & Enhanced explainability and human-AI collaboration & Improved transparency and user-aligned operation \\
\hline
Limited onboard computational resources for AI models & Using compact, finely tuned LLMs with efficient visual detection and depth estimation pipelines & High-level decision-making on resource-constrained UAVs & Widely deployable AI-powered UAV platforms \\
\hline
Complex environment perception and context-aware decision-making & Integrating domain-guided prompt execution with lightweight vision modules & Safe, context-aware decisions in dynamic environments & Intelligent operation in complex scenarios \\
\hline
Human intent understanding and mission execution & Systems combining language reasoning with visual perception and human-centric interaction & Understanding human intent and complex environments & More intelligent, interpretable, and reliable systems \\
\hline
\end{tabular}
\end{table*}

\begin{figure}[t!]
  \centering
  \includegraphics[width=0.45\textwidth]{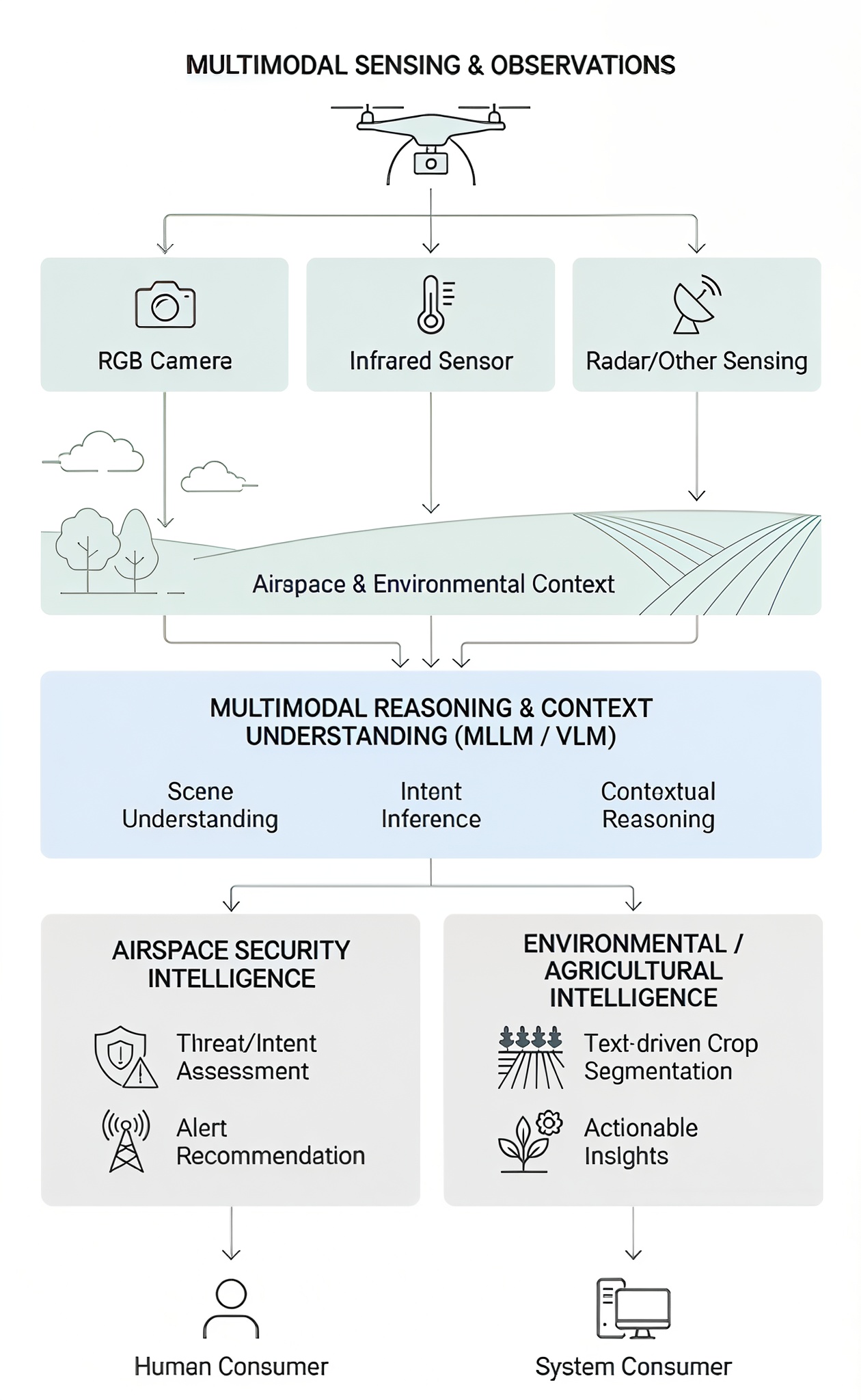} 
  \caption{Multimodal LLM/VLM framework for semantic understanding and analysis in UAV applications, integrating diverse sensor data for scene reasoning, threat assessment, and domain-specific insights.}
  \label{fig:semantic}
\end{figure}

\subsection{Semantic Understanding, Analysis, and Interpretation}

As multimodal large models gain traction in UAV-related domains, recent research has begun to explore their potential beyond navigation and control, extending into higher-level reasoning for security and remote sensing applications. Fig. \ref{fig:semantic} visualizes LLM/VLM-assisted semantic Understanding, Analysis, and Interpretation. Lei \textit{et al.}~\cite{lei2025enhancing} exemplify this shift in low-altitude airspace security by proposing a system that integrates MLLMs with heterogeneous perception data to infer the intent of non-cooperative UAVs. By fusing inputs from RGB cameras, infrared sensors, and radar, the system constructs structured representations, which are processed by an MLLM that leverages contextual knowledge and generative reasoning to assess adversarial behavior and recommend defensive actions. A confrontation scenario demonstrates the feasibility of this approach. The authors also identify key challenges – including robust communication, scalability to multi-UAV scenarios, and effective edge–cloud collaboration – that outline future directions for intelligent and adaptive airspace defense architectures.

\par
Extending multimodal models to environmental monitoring, Bie \textit{et al.}~\cite{11136213} investigate the use of VLMs for crop segmentation in UAV-assisted remote sensing. Their approach reframes segmentation as a text-driven coordinate prediction task by converting ground-truth masks into polygon-based XML representations. This formulation enables VLMs, fine-tuned with parameter-efficient LoRA techniques, to generate segmentation outputs directly through language. Experimental results show that Qwen2-VL-7B achieves the best performance among the evaluated models, particularly for tobacco segmentation, yielding competitive F1 scores compared with specialized vision-based methods. Although overall accuracy remains lower than that of task-specific architectures, the proposed method offers significant advantages in flexibility and generalization, positioning VLMs as promising tools for scalable and versatile agricultural remote sensing pipelines.

\par
Overall, multimodal AI is rapidly expanding UAV capabilities in both security and environmental monitoring. By integrating LLMs with visual and sensor data, UAVs can interpret complex scenes, infer intent, and make context-aware decisions in low-altitude airspace. At the same time, VLMs facilitate flexible environmental analysis, such as text-driven crop segmentation for precision agriculture. Together, these developments highlight the potential of multimodal AI to enable intelligent, adaptive, and generalizable UAV operations.

\subsection{Key Findings and Insights}

This section consolidates key observations and empirical findings on the deployment of Multimodal Large Language Models (MLLMs) in UAV applications. Emphasis is placed on quantitative evidence, benchmark results, and UAV-specific limitations that directly affect real-world reliability, efficiency, and safety as a result.

\subsubsection{\textbf{Benchmarking MLLMs for Multi-UAV Collaborative Perception}}

AirCopBench \cite{zha2025aircopbench} is the first benchmark explicitly designed to evaluate MLLMs in multi-UAV collaborative perception under degraded sensing conditions. The benchmark comprises over 2,900 multi-view aerial images and more than 14,600 VQA samples, spanning four task categories (Scene Understanding, Object Understanding, Perception Assessment, and Collaborative Decision) and 14 task types. Evaluation of 40 state-of-the-art MLLMs reveals a substantial performance gap: the best-performing model achieves only 59.17\% accuracy, lagging human performance by more than 24\% on average, with large variance across task categories. Fine-tuning results indicate that sim-to-real transfer can partially mitigate these gaps, highlighting the importance of UAV-specific data and training strategies for collaborative aerial perception.

\subsubsection{\textbf{Speculative Decoding for Efficient LLM Inference}}

Inference latency remains a critical bottleneck for UAV deployment, as conventional autoregressive decoding requires one forward pass per generated token. Speculative decoding improves efficiency by allowing multiple candidate tokens to be generated and verified in parallel within a single forward pass, significantly reducing end-to-end latency while preserving output correctness. Recent approaches, such as EAGLE-3, further eliminate the need for a separate draft model by leveraging the target model's internal hidden states. Although primarily evaluated in general-purpose LLM settings, these techniques are highly relevant to UAV platforms, where real-time responsiveness and energy efficiency are tightly constrained.

\subsubsection{\textbf{Multi-Agent Framework for MLLMs Training}}

The high computational cost of end-to-end multimodal training motivates modular and multi-agent alternatives. BEMYEYES \cite{huang2025my} adopts a collaborative framework in which a lightweight visual “perceiver” extracts scene information and a powerful text-only LLM acts as a “reasoner.” Through multi-turn interaction, this system enables multimodal reasoning without training large-scale vision–language models. Experimental results show that this fully open-source framework can match or surpass proprietary models such as GPT-4o on several knowledge-intensive tasks, demonstrating a scalable and resource-efficient path for multimodal reasoning in UAV systems.

\subsubsection{\textbf{Balancing Efficiency and Reasoning Diversity in Vision-Language Models for UAV Applications}}

Results in \cite{guan2025uav} comprising 50,019 samples organized into progressively challenging stages, further highlight that domain-specific data can outweigh sheer model scale, outperforming the posed UAV-VL-R1 model, which achieves strong reasoning performance while maintaining a compact footprint suitable for onboard deployment. Ablation studies reveal a key trade-off: while supervised fine-tuning (SFT) accelerates convergence and stabilizes early performance, excessive reliance on SFT reduces reasoning diversity, particularly in counting and numerical inference tasks. In contrast, the multi-stage GRPO framework enhances reasoning diversity and cross-task generalization by comparing intra-group rewards. The associated HRVQA-VL dataset, consisting of 50,019 samples organized into progressively challenging stages, further highlights that domain-specific data can outweigh sheer model scale, outperforming both 2B and 72B Qwen2-VL baselines under UAV constraints.

\subsubsection{\textbf{Understanding and Evaluating Hallucination Behavior in Aerial LVLMs}}

Basak et al. \cite{basak2025aerial} introduce AeroCaps, the first aerial-view image captioning dataset, comprising 1,256 human-annotated images. Their analysis shows that aerial perspectives significantly increase hallucination rates, especially for object identity, spatial relations, and numerical descriptions, with error frequency increasing with caption length. Notably, mitigation strategies such as object hints and LLM-based self-evaluation are only partially effective, underscoring the difficulty of reliable hallucination detection. These findings indicate that current LVLMs are insufficiently robust for real-world UAV deployment without further advances in dataset scale, rigorous evaluation, and grounding-aware model design.
The study emphasizes that, despite impressive performance on conventional benchmarks, current LVLMs are not yet sufficiently robust or trustworthy for deployment in real-world aerial applications, including automated surveillance, environmental monitoring, and disaster response. The results suggest that progress in this area will require not only improved model architectures and training strategies but also domain-specific datasets, rigorous evaluation protocols, and transparent reporting of failure modes. As aerial data becomes increasingly central to AI-driven decision-making, addressing these challenges will be critical to ensuring the safe deployment of AI.

\section{Ethical Challenges and Future Directions} \label{sec6}

This section reviews representative ethical challenges and future directions for secure, transparent, and accountable deployment in real-world UAV scenarios.
\par

\subsection{Securing UAV Operations: Privacy and Data Protection}

While LLMs enhance autonomy and reasoning in UAV networks, they also introduce security and privacy vulnerabilities, including prompt manipulation, leakage of sensitive sensor data, unauthorized access, and unsafe decision-making under adversarial conditions. When LLMs influence mission planning, resource scheduling, or control actions, these vulnerabilities can lead to severe operational failures.
LLM security encompasses practices that protect models, data, and outputs across development, deployment, and real-time operation. Core objectives include preventing unauthorized access, mitigating adversarial attacks such as jailbreaking and data poisoning, and ensuring compliance with safety and operational constraints.
\par
Jailbreaking is a particularly critical threat in LAUS. These attacks manipulate prompts to bypass safety constraints and induce unauthorized behavior. White-box attacks assume access to internal model information, enabling gradient- or logit-based manipulation, while black-box attacks rely solely on crafted prompts and are therefore more practical in real-world deployments.
Among black-box methods, template-completion attacks are particularly dangerous. By embedding harmful intent within structured templates, fictional scenarios, or role-playing contexts, attackers exploit LLMs’ pattern-completion behavior while evading keyword-based filters. Context-based manipulation and code injection further amplify these risks, particularly for models relying heavily on contextual coherence and demonstration-based reasoning.
These risks are amplified in UAV networks using ICL for real-time decision-making. For example, malicious demonstrations injected into scheduling prompts can bias UAVs toward suboptimal sensing actions, such as prioritizing sensors with poor channel conditions, thereby increasing packet loss, degrading network performance, and compromising mission objectives \cite{emami2025llm}. This illustrates how prompt-level attacks can propagate into system-level failures.

\subsection{Bias-Aware LLMs for Safer UAV Operations and Communications}

Bias in LLMs arises from imbalanced or unrepresentative training data and can directly affect safety and mission effectiveness in UAV systems, for example, through skewed threat assessment or unfair scheduling decisions. These risks are particularly severe in post-disaster scenarios characterized by uncertainty, limited communication, and strict latency constraints.
In large-scale SAR missions, relying on a single monolithic LLM for perception, planning, and scheduling can create an operational bottleneck. Under high cognitive load, such models are more prone to hallucinations, biased prioritization, and delayed responses, violating key requirements for multi-UAV operations, including real-time responsiveness, efficient resource use, and robustness to partial failures.
\par
To address these limitations, recent LLM-assisted UAV architectures distribute reasoning across ensembles of specialized agents responsible for distinct tasks, such as trajectory planning, sensing scheduling, or risk assessment. Bias can be further mitigated by supervisory LLMs that review decisions for fairness, consistency, and compliance before execution \cite{11108428}. This decentralized design improves situational awareness, reduces redundant communication, and enhances system integrity through cross-agent verification.

\subsection{Ethical LLMs in Action: Ensuring Accountability and Transparency for UAV Operation and Communication}

The deployment of LLMs in critical domains exposes a fundamental tension between capability and accountability. LLMs often function as opaque black boxes, making it difficult to interpret decisions, trace failures, or ensure consistent behavior. This opacity is exacerbated by the proprietary nature of many state-of-the-art models, limiting access to architectural and training details.
Enhancing causal reasoning offers a promising pathway to improved interpretability. By modeling cause-and-effect relationships rather than relying solely on pattern recognition, LLMs can provide more transparent and auditable decision logic, enabling error diagnosis and accountability in complex, real-world UAV tasks.
\par
This principle is exemplified by frameworks such as FlyAdapt, which integrates an LLM-assisted causal reasoning engine with predictive edge caching and hybrid verification for multi-UAV systems. By prioritizing interpretability and real-time validation, FlyAdapt achieves substantial latency reduction while intercepting the majority of diverse attacks, demonstrating how a transparent, modular design can support reliable deployment in high-stakes environments.
To manage residual risks, LLM observability must be adopted as a foundational practice. Observability connects input quality to output accountability through continuous monitoring, automated evaluation, and human feedback across the AI lifecycle. These mechanisms mitigate hallucination and performance drift, transforming LLMs from opaque components into governable systems.
\par
In summary, accountable LLM deployment requires advances in intrinsic interpretability through causal reasoning, transparent system architectures embedding verification, and rigorous observability for continuous governance. Together, these approaches enable secure and trustworthy integration of LLMs into UAV systems \cite{11202172}.

\subsection{Sustainable Intelligence: Environmental Concerns in LLMs}

The environmental footprint of LLMs presents a growing ethical challenge that requires a lifecycle perspective encompassing training, inference, deployment, and hardware production.
\par
From an energy perspective, training dominates carbon emissions, with large-scale models consuming electricity comparable to that of hundreds of households, whereas inference incurs lower per-query costs. However, at the scale of billions of daily interactions, inference energy consumption becomes a significant contributor to global emissions.
Environmental impacts are further amplified by lifecycle factors, including resource-intensive hardware manufacturing, embodied carbon in AI accelerators, and regional variation in data-center energy mixes. Design trade-offs also matter: larger models may be more efficient per high-quality output, whereas smaller models can be preferable under high request volumes.
\par
Importantly, sustainable trajectories remain achievable through mitigation strategies including model quantization, energy-efficient accelerators, deployment on low-carbon infrastructure, and standardized lifecycle environmental reporting. Aligning model design, infrastructure, and governance with transparency and renewable energy use is essential for reducing long-term ecological impact.
In summary, while per-query emissions remain modest, the aggregate and systemic environmental impacts of LLMs are substantial. Achieving sustainable intelligence requires coordinated optimization across computation, energy sourcing, hardware lifecycles, and transparent environmental governance \cite{wu2025unveiling},\cite{Grosser_2025}.

\subsection{Small and Edge-Deployable LLMs for UAV Applications}

The growing viability of small LLMs offers a promising research direction for UAV applications. These compact models can run efficiently on CPUs or entirely on-device, significantly reducing inference costs, enhancing privacy, and enabling deployment in offline or secure environments. Techniques such as QLoRA, model distillation, and FlashAttention have improved the performance of small models, allowing them to match or surpass larger LLMs on targeted tasks while remaining lightweight and practical for real-world workloads.
\par
Future research should investigate the development of domain-specific, task-oriented LLMs optimized for UAV networks that achieve high accuracy, reliability, and efficient resource use. Platforms such as Google's AI Edge Gallery demonstrate the potential for running advanced LLMs locally on devices, supporting offline execution, low latency, and privacy-preserving operations \cite{ahmed2025googlesAIEdgeGallery}. Leveraging such edge-deployable LLMs could enable real-time onboard intelligence for flight control, mission planning, and sensor data processing. Further research should also examine sustainable AI practices that balance performance with environmental impact, as well as scalable deployment strategies to maximize ROI, accessibility, and operational efficiency in UAV and other resource-constrained domains.

\subsection{Reducing Hallucinations Towards Reliable LLM Systems}

Addressing hallucinations in LLMs represents a critical area for future research. Hallucinations occur when models confidently generate plausible but incorrect information and can be classified as extrinsic – misrecalling facts present in training data – or intrinsic – contradicting or exceeding the input context \cite{hajji2025map}. These errors often result from ambiguous prompts, noisy or limited training data, restricted context windows, and the model's tendency to produce answers regardless of certainty. High-stakes domains, such as telecommunications, are particularly vulnerable to the consequences of hallucinations.
Future research should explore strategies to mitigate these errors, such as grounding models in verified data sources, refining prompt design, decomposing complex tasks, leveraging domain-specific models, and integrating human oversight. Advances in training methodologies, data integration, and evaluation frameworks could further enhance model reliability. Progress in these areas will be essential for developing trustworthy LLMs that consistently align with real-world knowledge and expectations.

A clear future research direction is the development of orchestration layers and system-level frameworks for LLMs. While current LLMs are already highly capable, their limitations in hallucination, reasoning, consistency, and context handling prevent their deployment in real-world applications with full reliability. Future research should focus on designing robust systems that manage, combine, and control LLMs, rather than simply scaling model size. Key areas include: a) Agent frameworks that enable LLMs to interact with tools, APIs, and other models in structured workflows, breaking complex tasks into verifiable subtasks; b) Reliability tooling, such as automated validators, fact-checkers, and monitoring systems, to detect and correct errors before outputs reach users;
c) Safety and ethical guardrails that prevent biased, harmful, or unsafe outputs through constraint mechanisms, moderation layers, and verification pipelines; d) Workflow automation that integrates HITL checkpoints and ensures repeatable, traceable, and efficient processes for high-stakes tasks.

\section{Conclusion}  
\label{sec9}

The integration of LLMs and MLLMs into UAV systems represents a significant step toward intelligent, adaptive aerial operations. This survey examined the convergence of LLM technologies with UAV communications, control, and ethical frameworks, providing a unified view of system architectures, methodologies, and real-world applications. By enabling advanced reasoning, natural language interaction, and context-aware decision-making, LLMs support applications ranging from real-time SAR to swarm coordination and mission planning. Techniques such as RAG and prompt engineering allow UAVs to operate dynamically in complex environments without extensive retraining, while MLLMs further extend these capabilities through multimodal perception and human-swarm interaction.
Despite their potential, LLM-assisted UAV systems pose critical challenges regarding bias, transparency, accountability, and environmental impact. This work is expected to accelerate the evolution of intelligent UAV ecosystems, enabling more reliable and responsible autonomous aerial systems for societal applications.

\bibliographystyle{IEEEtran}
\bibliography{references}
\end{document}